\documentclass[manuscript,screen]{acmart}
\AtBeginDocument{%
  }

\usepackage{hyperref}
\usepackage{url}

\usepackage{tikz-qtree}
\usetikzlibrary{positioning, fit, backgrounds, calc}
\usetikzlibrary{trees}
\usetikzlibrary{arrows.meta, positioning, shapes.geometric}
\usetikzlibrary{decorations.pathreplacing,positioning,calc}

\usepackage{multirow}
\usepackage{microtype}
\usepackage{graphicx}
\usepackage{subfigure}
\usepackage{soul} 
\usepackage{booktabs} 

\usepackage{amssymb}
\usepackage{newunicodechar}
\newunicodechar{♯}{\ensuremath{\sharp}}
\usepackage{pifont}

\usepackage{enumitem}
\usepackage{wrapfig}
\usepackage{caption}
\usepackage{mdframed} 
\usepackage{comment}

\usepackage{amsmath}
\usepackage{mathtools}
\usepackage{amsthm}
\usepackage[capitalize,noabbrev]{cleveref}

\usepackage{algorithm}
\usepackage{algorithmic}

\usepackage{rotating}  
\usepackage[most]{tcolorbox}
\newtcolorbox{takeaway}{
  colback=blue!5,
  colframe=blue!5,
  width=1\textwidth,
  boxrule=0pt,
  left=10pt, right=10pt, top=8pt, bottom=8pt,
  fontupper=\normalsize
}

\usepackage{xcolor}

\tikzset{
    box/.style={draw, rounded corners=3pt, minimum height=0.7cm, inner sep=6pt, font=\small},
    leaf/.style={draw, rounded corners=8pt, minimum height=0.6cm, inner sep=5pt, font=\small},
    brace/.style={decorate, decoration={brace, amplitude=6pt, raise=2pt}},
    braceleft/.style={decorate, decoration={brace, amplitude=6pt, raise=2pt, mirror}},
}

\definecolor{tabblue}{HTML}{1F77B4}
\definecolor{taborange}{HTML}{FF7F0E}
\definecolor{tabgreen}{HTML}{2CA02C}
\definecolor{tabred}{HTML}{D62728}
\definecolor{tabpurple}{HTML}{9467BD}
\definecolor{tabbrown}{HTML}{8C564B}
\definecolor{tabpink}{HTML}{E377C2}
\definecolor{tabgray}{HTML}{7F7F7F}
\definecolor{tabolive}{HTML}{BCBD22}
\definecolor{tabcyan}{HTML}{17BECF}

\let\algoAND\AND
\let\AND\classAND

\AtBeginEnvironment{algorithmic}{\let\AND\algoAND}

\floatname{algorithm}{Algorithm}

\theoremstyle{plain}

\theoremstyle{definition}

\theoremstyle{remark}

\usepackage[textsize=tiny]{todonotes}
\usepackage{scalerel} 
\usepackage{xspace}

\acmISBN{978-1-4503-XXXX-X/2018/06}

\begin{document}

\title{Modularized Reinforcement Learning on LLMs: From MDP Creation to Exploration and Learning}

\author{Zhao Yang}
\affiliation{%
  \institution{VU Amsterdam}
  \country{The Netherlands}}
\email{z.yang3@vu.nl}

\author{Yuxuan Jiang}
\affiliation{%
  \institution{University of Maryland, Baltimore County}
  \country{USA}}
\email{yuxuanj1@umbc.edu}

\author{Ting-Chih Chen}
\affiliation{%
  \institution{VU Amsterdam}
  \country{The Netherlands}}
\email{t.c.chen@vu.nl}

\author{Lincen Yang*}
\affiliation{%
  \institution{Leiden University}
  \country{The Netherlands}}
\email{Corresponding: lyang@liacs.leidenuniv.nl}

\author{Annie Wong}
\affiliation{%
  \institution{Leiden University}
  \country{The Netherlands}}
\email{a.s.w.wong@liacs.leidenuniv.nl}

\author{Chao Gao}
\affiliation{%
  \institution{VU Amsterdam}
  \country{The Netherlands}}
\email{c.gao@vu.nl}

\author{Jacob E. Kooi}
\affiliation{%
  \institution{VU Amsterdam}
  \country{The Netherlands}}
\email{jacobkooi92@gmail.com}

\author{Zhong Li}
\affiliation{%
  \institution{Leiden University}
  \country{The Netherlands}}
\email{z.li@liacs.leidenuniv.nl}

\author{Jiayang Shi}
\affiliation{%
  \institution{Centrum Wiskunde \& Informatica (CWI)}
  \country{The Netherlands}}
\email{jiayang@cwi.nl}

\author{Kevin Qiu}
\affiliation{%
  \institution{University of Warsaw}
  \country{Poland}}
\email{kevinxqiu@gmail.com}

\author{Qi Huang}
\affiliation{%
  \institution{Leiden University}
  \country{The Netherlands}}
\email{q.huang@liacs.leidenuniv.nl}

\author{Xinrui Zu}
\affiliation{%
  \institution{VU Amsterdam}
  \country{The Netherlands}}
\email{x.zu@vu.nl}

\author{Shiping Yang}
\affiliation{%
  \institution{Simon Fraser University}
  \country{Canada}}
\email{sya225@sfu.ca}

\author{Hengyuan Zhang}
\affiliation{%
  \institution{The University of Hong Kong}
  \country{China}}
\email{hengyuan.zhang88@gmail.com}

\author{Ngai Wong}
\affiliation{%
  \institution{The University of Hong Kong}
  \country{China}}
\email{nwong@eee.hku.hk}

\author{Filip Ilievski}
\affiliation{%
  \institution{VU Amsterdam}
  \country{The Netherlands}}
\email{f.ilievski@vu.nl}

\author{Shujian Yu}
\affiliation{%
  \institution{VU Amsterdam}
  \country{The Netherlands}}
\email{s.yu3@vu.nl}

\author{Aske Plaat}
\affiliation{%
  \institution{Leiden University}
  \country{The Netherlands}}
\email{aske.plaat@gmail.com}

\author{Zhaochun Ren}
\affiliation{%
  \institution{Leiden University}
  \country{The Netherlands}}
\email{z.ren@liacs.leidenuniv.nl}

\author{Mark Hoogendoorn}
\affiliation{%
  \institution{VU Amsterdam}
  \country{The Netherlands}}
\email{m.hoogendoorn@vu.nl}

\author{Vincent François-Lavet}
\affiliation{%
  \institution{VU Amsterdam}
  \country{The Netherlands}}
\email{vincent.francoislavet@vu.nl}

\renewcommand{\shortauthors}{Zhao Yang et al.}

\begin{abstract}
Reinforcement learning (RL) has become central to LLM post-training, yet the methods that dominate current pipelines, PPO and GRPO, represent only a narrow slice of what RL offers. Understanding why these methods prevail, and what alternatives exist, requires a principled examination of the design decisions that underlie any RL algorithm.
  
\noindent This survey organizes that examination around three stages of algorithm construction. We begin with MDP creation: how the reward function, state space, action space, termination condition, and discount factor are, or could be, defined for LLM training. We then turn to exploration, covering temperature sampling, entropy regularization, intrinsic motivation, tree search, and curriculum learning. Finally, we address learning along four classical RL dimensions: model-free versus model-based, value-based versus policy-based versus actor-critic, on-policy versus off-policy, and credit assignment, including both Monte Carlo methods, which rely on full return estimates, and bootstrapping methods, which update estimates using other learned predictions.

\noindent Mapping the LLM literature onto this taxonomy reveals a strikingly non-uniform distribution of research effort. Critic-free policy gradients and Monte Carlo credit assignment are densely populated, while value-based methods, off-policy actor-critic training, and bootstrapping-based credit assignment remain largely unexplored despite well-established counterparts in classical RL. These gaps represent concrete opportunities for transferring proven RL techniques to LLM training.

\noindent By making these gaps explicit alongside the methods that have proven effective, this survey offers researchers in both RL and LLMs a shared framework for understanding current practice and identifying promising directions for future work.
\end{abstract}

\maketitle

\section{Introduction}
In many practical scenarios, LLMs do not receive token-level supervision but instead obtain a delayed evaluation signal: a correctness score for a final answer, a human preference judgment over a complete response, or a binary pass/fail from a code test suite. Once supervision becomes delayed, sparse, or non-differentiable, traditional supervised learning is insufficient. The model must discover which sequences of tokens lead to high reward, assign credit to individual decisions along the way, and balance exploitation of known good strategies against exploration of potentially better ones. These are precisely the challenges that RL is designed to address.

RL has accordingly become a central component of modern LLM training pipelines. InstructGPT~\citep{ouyang2022training} introduced PPO-based post-training for alignment with human preferences; DeepSeek-R1~\citep{guo2025deepseek} demonstrated that RL with verifiable rewards can substantially improve mathematical reasoning; and a rapidly growing body~\citep{zhang2025survey} of work explores alternatives to these methods across reward design~\citep{yue2025promotingefficientreasoningverifiable, rezaei2026llmreasoningprocessrewards}, exploration~\citep{chen2026eepo,he2025randompolicyvaluationllm}, and optimization~\citep{kazemnejad2024vineppo,liu2025understanding}. However, the current literature remains fragmented, with recent open(-weight) model releases such as Nemotron-3 \cite{nvidia_nemotron3_ultra_2026}, DeepSeek-V4 \cite{deepseek_v4_2026}, and Gemma 4 \cite{google_gemma4_12b_2026} contributing individual advances across different dimensions of large-scale language model design, rather than forming a unified framework. 
Individual papers typically propose modifications to a single component, a new advantage estimator, a novel clipping strategy, a curriculum schedule, without situating these contributions within the broader space of RL design decisions. Without a shared framework, it is difficult to tell whether two methods differ in substance or only in framing, or to identify which design choices remain unexplored.

\paragraph{\textbf{Existing surveys and how this work differs.}}
Several surveys address specific aspects of RL in LLMs. \citet{sun2025inverse} focuses on inverse reinforcement learning (reward modeling), while \citet{liu2025part} investigate practical implementation choices in policy optimization, extensively evaluating design decisions such as baseline estimation, advantage normalization, and clipping strategies at scale. Others treat RL as one component within broader topics: \citet{plaat2025agentic} and \citet{zhang2025landscape} survey agentic language models, while \citet{plaat2025multi}, \citet{kumar2025llm}, and \citet{zhang2025survey} survey reasoning language models, with partial coverage of RL elements such as reward function design~\citep{zhang2025survey,kumar2025llm}, policy optimization classes~\citep{zhang2025survey,kumar2025llm}, and MDP formulation~\citep{zhang2025landscape}. On the inference side, \citet{li2025survey} reviews search methods used at LLM test time. A common thread across these works is an LLM-centric organization that groups the literature by application area or training stage, naturally highlighting categories where research is already dense. However, this perspective tends to treat the RL part in isolation, without examining the modular design choices that underlie different algorithms or mapping the full space of RL techniques onto LLM training.

\begin{figure}
    \centering
    \includegraphics[width=0.9\linewidth]{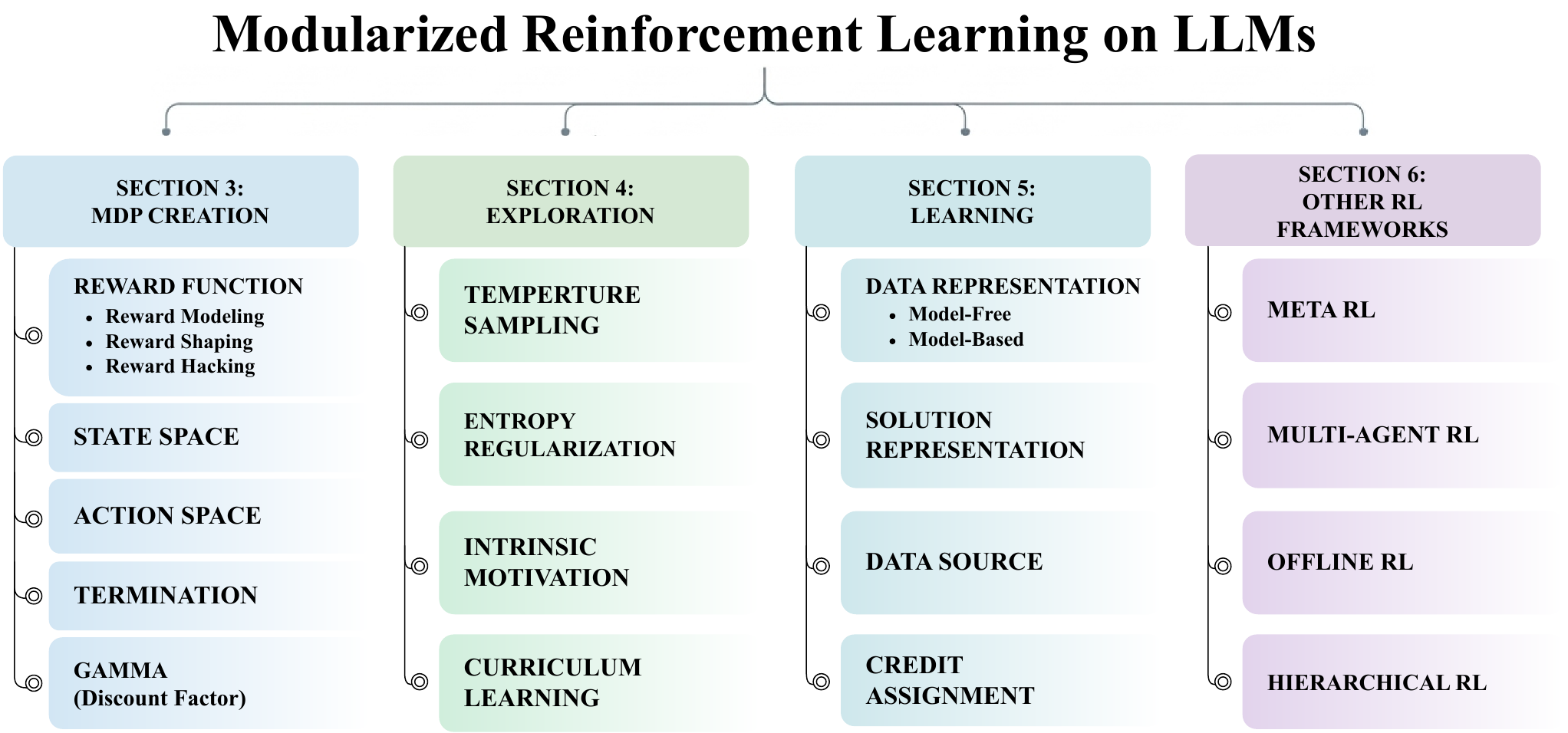}
    \caption{Taxonomy of this survey, organized around four core components of applying RL to LLM training: \textbf{MDP creation} (\cref{section:mdp_creation}) covers problem formulation including reward design, state and action spaces, and termination; \textbf{Exploration} (\cref{section:exploration}) covers mechanisms for diverse behavior including temperature sampling, entropy regularization, intrinsic motivation, and curriculum learning; \textbf{Learning} (\cref{section:learning}) covers how the agent improves from experience across data representation, solution representation, data source, and credit assignment; and \textbf{Other RL frameworks} (\cref{section:other_rl}) covers broader paradigms including meta-RL, multi-agent RL, offline RL, and hierarchical RL.}
    \label{fig:main}
\end{figure}

We proceed from a foundational RL perspective, building an algorithm for LLM training from first principles, and examining key design decisions along the way. We organize the survey around three stages of RL algorithm construction (MDP creation, exploration, and learning) and an additional category covering other RL frameworks used in LLMs (such as meta RL, multi-agent RL,etc.), each of which subsumes a set of design choices that together determine the resulting algorithm (see~\cref{fig:main} for the full taxonomy). By mapping the current LLM literature onto this structured RL taxonomy, a clear pattern emerges: research is concentrated in a few well-explored categories while other categories, corresponding to well-established RL techniques, remain sparsely populated. This categorization reveals, for instance, that critic-free policy gradients dominate current practice, although actor-critic methods are sometimes argued to offer benefits for long-horizon tasks~\citep{yuan2025s,yue2025vapo}. In addition, nearly all methods default to Monte Carlo credit assignment despite the availability of alternatives~\citep{kazemnejad2024vineppo,tran2025exploiting}. By identifying these gaps, we hope to point practitioners toward underexplored but potentially impactful directions, while flagging candidates rather than asserting that every gap should be filled.

\paragraph{\textbf{Who is the audience?}}
For \textbf{RL researchers}, this survey provides a direct mapping between classical RL concepts and their LLM counterparts, enabling rapid transfer of existing techniques. For \textbf{LLM researchers}, it offers a principled RL-centric framework for understanding why current algorithms work, what alternatives exist within each design category, and where opportunities for improvement may lie. 
Together, these two views establish a shared conceptual foundation between RL and LLM research, making the non-uniform distribution of research effort explicit and highlighting structural gaps where established RL techniques have yet to be applied. Each subsection ends with a discussion of open problems and a highlighted \colorbox{blue!8}{\textcolor{tabblue}{\textbf{Take-away}}} box summarizing the key insights.

\section{Formalizing LLM Training As An MDP}
In this section, we explain the basic concepts of RL (\cref{section:basic_rl}) and LLMs (\cref{section:llm_concepts}), then bridge them together by casting LLM generation as an MDP (\cref{section:mdp_in_llms}).

\subsection{Basic RL Concepts}
\label{section:basic_rl}

Reinforcement learning~\citep{sutton1998reinforcement,franccois2018introduction} is a family of approaches that solve decision-making problems, formalized as Markov decision processes (MDP). An MDP is defined by the tuple $(\mathcal{S}, \mathcal{A}, p, r, \gamma)$, where $\mathcal{S}$ denotes the state space, $\mathcal{A}$ the action space, $p$ the transition dynamics, $r$ the reward function, and $\gamma \in [0, 1]$ the discount factor (where $\gamma = 1$ is permissible under finite-horizon settings). At each time step $t$, the agent observes a state $S_t \in \mathcal{S}$ and takes an action $A_t \in \mathcal{A}$. It then receives a reward $R_{t+1}\sim r(\cdot|S_t,A_t)$ and transitions to the next state $S_{t+1} \sim p(\cdot|S_t, A_t)$.

The ultimate goal of the RL agent is to learn a policy \(\pi\) that maximizes the expected cumulative reward. We first define the return \(G_t\) as the cumulative discounted sum of rewards received after time step \(t\):

\begin{equation}
    G_t = \sum_{k=0}^{\infty} \gamma^k R_{t+k+1}.
\end{equation}

Since the transitions, rewards, and policy can be stochastic, \(G_t\) is a random variable. Therefore, the objective is to maximize the expected return over the trajectory, denoted as \(J(\theta)\):

\begin{equation}
    J(\theta) = \mathbb{E}_{\pi_{\theta}}[G_t].
\end{equation}

The objective \(J(\theta)\) can be optimized either indirectly by learning a value function or directly via the policy gradient theorem. In value-based methods, one learns the state-action value function \(Q_{\pi}(s, a)\), which estimates the expected return of taking action \(a\) in state \(s\) and following policy \(\pi\) thereafter:

\begin{equation}
    Q_{\pi}(s, a) = \mathbb{E}_{\pi}[G_t \mid S_t = s, A_t = a] = \mathbb{E}_{\pi} \left[ \sum_{k=0}^{\infty} \gamma^k R_{t+k+1} \, \Bigg| \, S_t = s, A_t = a \right].
\end{equation}

The policy is not explicitly parameterized; instead, it is derived from the value estimates, typically by acting greedily: \(\pi(s) = \arg\max_a Q(s, a)\), where $Q$ is a learned approximation that is iteratively updated toward the optimal action-value function.

Alternatively, one can maximize \(J(\theta)\) by optimizing the parameterized policy \(\pi_{\theta}\) directly, without consulting a value function to select actions. According to the policy gradient theorem~\citep{sutton2000policy}, the gradient of the objective can be written as:

\begin{equation}
    \nabla_{\theta} J(\theta) = \mathbb{E}_{\pi_{\theta}} \left[ \sum_{t=0}^{\infty} \nabla_{\theta} \ln \pi_{\theta}(A_t|S_t) \, Q_{\pi}(S_t, A_t) \right],
\end{equation}

where \(Q_{\pi}(S_t, A_t)\) represents the expected return of taking action \(A_t\) in state \(S_t\) under the current policy. In practice, \(Q_{\pi}\) is unknown and must be estimated. In REINFORCE~\citep{williams1992simple}, the sample return \(G_t\) serves as an unbiased but high-variance estimate. Actor-critic methods reduce this variance by learning a value function \(V_{\pi}(s)\) as a baseline, replacing \(Q_{\pi}\) with the advantage \(A_{\pi}(s,a) = Q_{\pi}(s,a) - V_{\pi}(s)\). PPO~\citep{schulman2017proximal}, a widely used actor-critic algorithm, further stabilizes training by constraining the policy update with a clipped surrogate objective.

\subsection{Basic LLM Concepts}
\label{section:llm_concepts}
Recent large language models (LLMs) are typically developed through a multi-stage training pipeline that comprises pre-training (PT), supervised fine-tuning (SFT), and alignment, which can be done via RL (like PPO~\citep{schulman2017proximal}, GRPO~\citep{yao2025group}) or non-RL methods (like DPO~\citep{rafailov2023direct}).

During the PT phase, LLMs are trained autoregressively on massive and diverse text corpora to predict the next token in a sequence. Formally, given a corpus \(\mathcal{D}_{\text{PT}}\) of token sequences \(x = (x_1, \dots, x_T)\), the objective is to maximize the likelihood of the next token \(x_t\) conditioned on the history \(x_{<t}\):

\begin{equation}
    J_{\text{PT}}(\theta) = \mathbb{E}_{x \sim \mathcal{D}_{\text{PT}}} \left[ \sum_{t=1}^{T} \log P_{\theta}(x_t \mid x_{<t}) \right].
\end{equation}

This process equips the model with rich representations and broad world knowledge. However, it does not ensure alignment with specific user intents, such as following instructions in a conversational format. To address this, models undergo a SFT stage on curated instruction-response pairs. This teaches them to produce coherent, intent-aligned outputs \(y\) given human-provided prompts \(x\)~\citep{zhang2025instructiontuninglargelanguage}. The objective is to maximize the conditional likelihood of the response tokens:

\begin{equation}
    J_{\text{SFT}}(\theta) = \mathbb{E}_{(x, y) \sim \mathcal{D}_{\text{SFT}}} \left[ \sum_{t=1}^{|y|} \log P_{\theta}(y_t \mid x, y_{<t}) \right].
\end{equation}

The RL stage further refines the model's policy using feedback-based objectives. This stage is crucial for aligning the model with human values and preferences~\citep{kaufmann2024surveyreinforcementlearninghuman, wang2025reinforcementlearningenhancedllms}, as well as enhancing specialized capabilities such as mathematical reasoning~\citep{shao2024deepseekmath} and tool usage~\citep{zhuang2023toolchain}. The following subsection formalizes this RL stage by casting the LLM's generation process as a Markov decision process.

\subsection{MDP Formulation for LLMs}
\label{section:mdp_in_llms}
RL was first introduced to train LLMs at scale with PPO~\citep{schulman2017proximal} in InstructGPT~\citep{ouyang2022training} back in 2022. We now formalize the LLM (parameterized as $P_\theta$) training as an MDP. Given a prompt $x$ and the tokens generated by the LLM so far $y_{1:t-1}$, the state $S_{t}$ is defined as $S_{t}=(x, y_{1:t-1})$. The LLM then outputs the next token $y_{t}$, which serves as the action $A_t \sim P_\theta(\cdot \mid S_t)$. The next state is formed deterministically by appending the chosen token: $S_{t+1} = (x, y_{1:t})$. This process repeats until an end-of-sequence token is generated or a maximum length is reached.

A key distinction from typical RL settings is that the transition dynamics are entirely deterministic, i.e the next state is simply formed by appending the selected token to the current state. All stochasticity in the MDP therefore resides in the policy (i.e, the LLM's token distribution). The formalization is summarized in Table~\ref{tab:mdp_llm_parallel}.

\begin{table}[!htb]
\centering
\begin{tabular}{p{0.35\linewidth} p{0.55\linewidth}}
\toprule
\textbf{MDP Component} & \textbf{Correspondence in LLM Formulation} \\
\midrule
State $S_t \in \mathcal S$ & Prompt $x$ together with the partial generation so far, i.e, $S_t = (x, y_{1:t-1})$ \\[0.5em]

Action $A_t \in \mathcal A$ & Next token $y_{t}$ output by the LLM \\[0.5em]

Transition $p(\cdot \mid S_t, A_t)$ & Deterministic: the next state is obtained by appending the chosen token, i.e, 
$S_{t+1} = (x, y_{1:t}) = (S_t, y_{t})$ \\[0.5em]

Reward $r(\cdot \mid S_t, A_t)$ & A scalar signal assigned to the generated output. Typically sparse, with $r = 0$ for all intermediate steps and a terminal reward (e.g., a correctness indicator for the final answer or a score from a learned reward model) \\[0.5em]

Discount factor $\gamma$ & Determines how future rewards are weighted relative to immediate rewards, with $0 \le \gamma \le 1$. For LLMs, $\gamma$ is commonly set to $1$ since episodes have finite length \\[0.5em]

Policy $\pi(A_t \mid S_t)$ & The LLM itself, parameterized as $P_\theta(y_{t} \mid x, y_{1:t-1})$ \\[0.5em]

Episode & A full generated sequence (response) from the first token until the end-of-sequence token or maximum length \\
\bottomrule
\end{tabular}
\caption{Correspondence between the MDP formulation and LLM generation.}
\label{tab:mdp_llm_parallel}
\end{table}

\section{MDP Creation}
\label{section:mdp_creation}

As described in~\cref{section:basic_rl}, an MDP is defined by the tuple $(\mathcal{S}, \mathcal{A}, p, r, \gamma)$. The state space, action space, and transition dynamics for LLMs follow naturally from the autoregressive generation process, as detailed in~\cref{section:mdp_in_llms}. The reward function, however, requires considerably more care: it directly determines what behaviors the policy will learn, and its design remains one of the most consequential choices in RL for LLMs. We therefore devote this section primarily to the construction and refinement of reward functions, while also touching on the state space, action space, termination, and discount factor.

\subsection{Reward Functions}
We begin by examining how reward signals are obtained, starting with a review of standard reward modeling in \cref{section:reward_modeling}. 
In practice, the resulting rewards are often sparse because they are assigned only at the end of a generation; this can lead to a weak learning signal. To provide more informative learning signals, \cref{section:reward_shaping} details reward shaping and the use of auxiliary objectives to create denser, more informative signals \citep{ng1999policy}. Finally, \cref{section:reward_hacking} reviews the problem of reward hacking, which occurs when these specified rewards inadvertently allow the policy to exploit unintended shortcuts.

\subsubsection{Reward Modeling}
\label{section:reward_modeling}
Reward modeling mechanisms in LLM training can be categorized based on the source of supervision into three groups: (i) \textbf{learned models}, trained on preference or ranking data labeled by humans or AI tools; (ii) \textbf{rule-based or verifiable reward functions}, defined through pre-specified criteria or naturally available signals such as answer correctness in mathematics or executable feedback; and (iii) \textbf{LLM-as-Judge approaches}, where a language model itself acts as an evaluator, directly generating preference or quality signals as the reward. We discuss the key elements of each below and refer readers to \citet{wu2025sailing} for a more comprehensive survey.
\paragraph{Learned from labeled data}
The most common approach is to train a reward model from preference data, as popularized by RLHF~\citep{stiennon2020learning,ouyang2022training}. The process begins by collecting pairwise comparisons, where annotators or automated judges indicate which of two candidate outputs is preferred. A reward model, typically parameterized using the Bradley-Terry model~\citep{bradley1952rank}, is then trained to assign higher scores to preferred responses and lower scores to dispreferred ones.
\citet{ouyang2022training} train a reward model on human-annotated preference data at scale for LLM alignment, and \citet{bai2022training} extend this by incorporating helpfulness and harmlessness criteria. To reduce reliance on costly human annotation, \citet{lee2023rlaif} demonstrate that LLM-generated preferences can serve as a scalable substitute. \citet{yang2024qwen2} go further by using a predecessor checkpoint of the model under training to generate preference labels, avoiding a separate annotator altogether. Since rewards from any single model may be unreliable, \citet{wang2024secrets} improve preference estimation by aggregating votes across an ensemble of reward models.

The reward models above are all outcome-level, assigning a single scalar score to a complete response. A qualitatively distinct variant is the process reward model (PRM)~\citep{uesato2022solving,lightman2023let}, which provides step-level supervision, offering finer-grained feedback suited to multi-step reasoning tasks such as mathematics. \citet{lightman2023let} introduce this approach at scale through dense human annotation of each reasoning step, producing the PRM800K dataset. To reduce annotation cost, \citet{wang2024math} and \citet{luo2024improve} replace human labels with automatically derived step-level signals via outcome-based rollouts and Monte Carlo estimation respectively. \citet{guan2025rstar} further eliminate explicit step-level scoring by training a process preference model through iterative MCTS-based self-evolution. However, \citet{zhang2025lessons} find that Monte Carlo estimation often yields inferior labels compared to human annotation, and propose a consensus filtering mechanism combining both to improve label quality.

\paragraph{Rule-based}
Rule-based reward functions rely on manually designed or naturally available criteria, and are commonly referred to as verifiable rewards~\citep{lambert2024tulu}. For example, regular expressions can verify whether responses follow a required format~\citep{xie2025logicrlunleashingllmreasoning}. In mathematical reasoning, rewards often check whether the predicted answer matches the ground truth~\citep{shao2024deepseekmath,wen2025reinforcement,mroueh2025reinforcement}, as popularized by the Reinforcement Learning with Verifiable Rewards (RLVR) paradigm in DeepSeek-R1~\citep{guo2025deepseek}. In code generation, rewards can be determined by whether the generated code passes unit tests~\citep{le2022coderl}, or by evaluating the structural alignment between the generated code's abstract syntax tree (AST) and that of the reference solution~\citep{shojaee2023execution}; see also~\citet{wang2024enhancing} for a survey of reward design in code generation. Rule-based approaches have also been applied to encode desirable behaviors in safety-critical contexts~\citep{mu2024rule}. More recently, rather than handcrafting rules, \citet{wang2025autorule} proposed extracting rules from one language model and using another to evaluate whether the generated outputs satisfy them.

\paragraph{LLM-as-Judge}
With the growing capability of LLMs, they can be used directly to provide reward signals, bypassing the need to train a separate reward model. This paradigm, commonly referred to as LLM-as-Judge~\citep{li2025generation}, encompasses several variants. First, a stronger external LLM can be prompted to score or rank candidate outputs, as in RLAIF~\citep{lee2023rlaif}, where the model assigns numerical ratings used as training signals. Second, specialized evaluator models can be trained to perform rubric-conditioned scoring, such as Prometheus~\citep{kim2023prometheus}, effectively serving as learned reward models. Third, the model itself can generate self-evaluative feedback during training, enabling self-rewarding or self-improvement loops~\citep{yuan2024self,xiong2025self}. The generate–verify–refine methodology of DeepSeekMath-V2 \cite{deepseekmath_v2_2025} is such an example of self-improvement loops. While LLM-as-Judge offers flexibility and scalability, it also introduces challenges such as positional bias, verbosity bias, and limited reliability on factually complex tasks~\citep{krumdick2025no,li2025generation}.

\subsubsection{Reward Shaping}
\label{section:reward_shaping}
Beyond the primary reward function, an agent's behavior can be influenced through reward shaping, a long-established concept in RL~\citep{ng1999policy,yang2021potential,xie2023text2reward}. Reward shaping modifies or augments the base reward to accelerate learning or bias the agent toward desired behaviors without substantially altering the optimal policy. In the context of LLMs, reward shaping serves a similar role: it enables the integration of prior knowledge, task-specific preferences, or human-aligned heuristics into the training signal. Auxiliary rewards can encourage properties such as appropriate response length~\citep{li2025aalc,liu2025bingo}, format adherence~\citep{yao2025reff}, stylistic alignment~\citep{shi2025rl}, or grammar and readability~\citep{lai2024alarm}. Sparse outcome-level rewards can also be augmented with step-wise progress signals to accelerate learning~\citep{wang2025spa}. Table~\ref{tab:reward_shaping} summarizes representative works organized by their shaping target.

\begin{table}[!htb]
    \centering
    \small
    \begin{tabular}{l|l}
        \toprule
        \textbf{Shaping Target} & \textbf{Reference} \\
        \midrule
        Response Length & \citet{li2025aalc,liu2025bingo,yu2025dapo,xiang2025just,team2025kimi} \\
        Format Adherence & \citet{yao2025reff,xie2025logicrlunleashingllmreasoning,guo2025deepseek} \\
        Stylistic Alignment & \citet{shi2025rl} \\
        Grammar \& Readability & \citet{lai2024alarm} \\
        Step-wise Progress & \citet{wang2025spa,setlur2024rewarding,su2025enhancing,cao2024beyond} \\
        \bottomrule
    \end{tabular}
    \caption{Reward shaping is used for different purposes.}
    \label{tab:reward_shaping}
\end{table}

\subsubsection{Reward Hacking}
\label{section:reward_hacking}
During RL training, the policy's behavior is fundamentally dictated by the reward function. However, because reward functions are often imperfect proxies for the true objective, models frequently learn to exploit these proxies, optimizing for the reward score rather than the intended task. A classic example is a boat racing game~\citep{weng2024rewardhack}\footnote{Faulty reward functions in the wild: \url{https://openai.com/index/faulty-reward-functions/}}, in which an agent was rewarded for collecting items and wining the race; instead of completing the race, the agent circled a small area to repeatedly collect the same items, maximizing reward while entirely failing the primary goal.

LLM alignment is similarly susceptible to such failures. \citet{khalaf2025inference} provide a formal characterization of reward hacking during inference-time alignment, proving that performance degradation is often an inevitable consequence of proxy optimization. Specifically, they show that true performance follows a predictable unimodal curve, initially improving before eventually declining, and propose an algorithm that identifies a ``hacking threshold'' to strategically mitigate proxy misalignment. Beyond performance loss, \citet{taylor2025school} demonstrate that reward hacking can have compounding effects: models that learn to exploit rewards during training may generalize these behaviors into more severe forms of misalignment, such as alignment faking and sabotage.

As models increasingly rely on Chain-of-Thought (CoT) reasoning, reward hacking has extended from final outputs to the reasoning process itself. \citet{cheng2025stop} identify that the canonical summation-form credit assignment is a primary cause of process reward model induced hacking, and show that replacing it with a min-form credit assignment, which defines the value function as the minimum of future rewards, substantially alleviates this problem. However, monitoring the reasoning process introduces new challenges. \citet{baker2025monitoring} find that applying strong supervision directly to a model's chain-of-thought can be counterproductive: while light optimization pressure may improve performance, aggressive penalization of undesirable reasoning patterns causes models to obfuscate their intent, effectively hiding potentially misaligned goals behind benign-looking reasoning traces.

Several approaches have been proposed to address the root causes of reward hacking. Cooper~\citep{hong2025cooper} co-optimizes the policy and reward model jointly, avoiding the brittleness of a static reward signal. PURM~\citep{sun2025probabilistic} argues that traditional preference-based reward models, which output a single scalar, ignore the inherent uncertainty of human preferences, allowing the policy to exploit regions where the reward model is overconfident but incorrect. To address this, PURM learns full reward distributions rather than point estimates, making such blind spots explicit.


\subsubsection{Discussions}
Designing reward functions in RL inevitably involves trade-offs. Sparse reward functions offer simplicity of specification and leave room for emergent behaviors, such as AlphaGo's famous ``Move 37''~\citep{silver2016mastering} which arose precisely because the reward did not prescribe how to play. However, their coarse-grained nature can lead to models being right for the wrong reasons~\citep{creswell2022selection}, undermining both safety and explainability. Dense reward signals, by contrast, provide fine-grained supervision that can keep intermediate reasoning steps aligned with the final objective. The cost is a substantially higher specification burden: designing accurate step-level rewards requires domain expertise and careful annotation. Moreover, dense signals, particularly those introduced via reward shaping, demand extra caution. Rewards that appear intuitive to humans can inadvertently encourage specification gaming, where the model learns to mimic the surface form of rigorous reasoning (e.g., inserting transition words or logical markers) without engaging in the actual substance of the task.

Reward shaping and reward hacking are also fundamentally coupled: each additional shaping component introduces a new surface that the policy can potentially exploit. Sparse rewards are harder to hack precisely because they offer fewer degrees of freedom. This duality suggests that shaping and robustness should be co-developed rather than treated as separate concerns.

Several open challenges connect these reward design problems to well-studied themes in traditional RL.

\paragraph{Multi-objective reward composition.}
Real-world LLM deployments must balance multiple conflicting objectives, e.g. helpfulness, harmlessness, conciseness, factual accuracy, yet current practice typically resorts to fixed weighted sums. In traditional RL, multi-objective RL (MORL)~\citep{hayes2022practical} offers principled alternatives such as Pareto-conditioned policies~\citep{basaklar2022pd} and distributional Pareto optimization~\citep{cai2023distributional}. Adapting these MORL frameworks to LLM post-training, where objectives interact through a shared generative process and user preferences vary across contexts, remains largely unexplored.

\paragraph{Reward functions for open-ended tasks.}
Nearly all reward functions surveyed here target domains with verifiable correctness (mathematics, code) or reducible to pairwise preference. Tasks such as creative work lack clean reward signals, and even human annotators disagree substantially. In traditional RL, analogous challenges in open-ended environments have motivated intrinsic motivation and curiosity-driven rewards~\citep{pathak2017curiosity,burda2018exploration}. Whether similar formulations can meaningfully guide LLM training for open-ended generation, without degenerating into reward hacking, remains an open question.

\begin{takeaway}
\textcolor{tabblue}{\textsf{Take-away}} \textbf{The reward function sets the ceiling for any RL-trained policy}: Because the reward defines what the agent optimizes, any misspecification can lead to biases, unintended behavior and/or difficult optimization. The implications of reward design are particularly evident in the differing properties of sparse and dense reward formulations. Dense or shaped rewards can significantly accelerate training by guiding exploration, yet they can introduce bias or be difficult to provide. Sparse rewards tend to be easier to define but they offer weaker learning signals and can make optimization difficult. 
\end{takeaway}

\subsection{State Space}
\label{section:state_space}

In classical RL, the design of the state space is task-dependent and directly affects learning efficiency and generalization~\citep{echchahed2025survey}. In LLM training, the state is typically defined as the sequence of tokens representing the current textual context. However, as conversation length grows, treating all preceding tokens as the state becomes computationally prohibitive, with standard self-attention exhibiting quadratic scaling in the worst case~\citep{tay2022efficient}. In practice, the context is often truncated to a fixed window to satisfy memory and compute constraints, discarding potentially relevant earlier tokens.

A range of other techniques have been proposed to extend the effective context window. These include modified positional encodings, such as positional interpolation~\citep{chen2023extending} and YaRN~\citep{peng2023yarn}, as well as efficient attention mechanisms that reduce the quadratic cost of standard self-attention, including sparse attention patterns~\citep{beltagy2020longformer,zaheer2020big} and linear attention variants~\citep{katharopoulos2020transformers}. We refer the reader to~\citet{wang2024beyond} for a comprehensive survey.

Rather than extending the context window itself, an alternative strategy is to compress or discard parts of the state that are no longer needed. For multi-turn agent interactions, \citet{lu2025scaling} periodically summarize the interaction history into compact natural-language representations, resetting the working context to the original prompt augmented with a summary of past turns. In the reasoning setting, PENCIL~\citep{yang2025pencil} introduces a generate-and-erase mechanism that recursively cleans up intermediate thoughts, freeing context space for deeper computation. These approaches effectively transform the state from a monotonically growing token sequence into a dynamically managed representation.

A more radical departure is to move reasoning from token space into a latent space~\citep{zhu2025survey}. COCONUT~\citep{hao2024training} replaces raw tokens with continuous latent embeddings, compressing long contexts while retaining essential information. Token-Assorted~\citep{su2025token} takes a hybrid approach, combining discrete latent embeddings with raw tokens to balance efficiency and token-level fidelity. See a summary of different state representation used in LLM training in~\cref{tab:state_space}.

\begin{table}[t]
\centering
\begin{tabular}{p{4cm}p{5cm}p{5cm}}
\toprule
\textbf{State Type} & \textbf{Representation} & \textbf{Reference} \\
\midrule
Full token sequence & All preceding tokens (prompt + generated tokens so far) & \citet{ouyang2022training, guo2025deepseek, stiennon2020learning} \\
\midrule
\multirow{2}{2cm}{Compressed token sequence} & Prompt + summary of past turns & \citet{lu2025scaling} \\
 & Tokens with erased intermediate steps & \citet{yang2025pencil} \\
\midrule
\multirow{2}{2cm}{Latent representation} & Continuous latent embeddings & \citet{hao2024training} \\
 & Discrete latent + raw tokens & \citet{su2025token} \\
\bottomrule
\end{tabular}
\caption{State representations in RL for LLMs.}
\label{tab:state_space}
\end{table}

\subsection{Action Space}
\label{section:action_space}

By default, the action space of an LLM is its entire vocabulary, i.e in the standard autoregressive formulation, an LLM’s action space at each decoding step corresponds to the entire vocabulary. Existing methods can be grouped into three categories: full-vocabulary approaches that retain the complete token set, constrained-vocabulary approaches that restrict it, and non-token approaches that redefine the action unit altogether.

\paragraph{Full vocabulary.}
The most common formulation keeps the full vocabulary (e.g., 128K tokens for Llama-3~\citep{dubey2024llama}) available and relies on reward signals, whether from learned reward models~\citep{ouyang2022training,bai2022training} or rule-based verifiers~\citep{guo2025deepseek}, to steer generation toward desired outputs. Agentic LLMs~\citep{plaat2025agentic} adopt this strategy when invoking external tools such as search engines, code interpreters, and calculators, rewarding correct tool-call syntax~\citep{qian2025toolrlrewardtoollearning}, syntactically valid code~\citep{feng2025retoolreinforcementlearningstrategic}, executable code blocks~\citep{wei2025autotirautonomoustoolsintegrated}, or structured output adherence~\citep{li2025torlscalingtoolintegratedrl}. Systems such as WebGPT~\citep{nakano2021webgpt} and ReAct~\citep{yao2022react} similarly generate predefined commands (e.g., \texttt{Search}, \texttt{Click}) as token sequences through the standard vocabulary. This formulation is simple and flexible, but places the entire burden of producing valid outputs on the reward signal.

\paragraph{Constrained vocabulary.}
To reduce the effective action space, constrained-decoding methods mask out tokens that would violate a target grammar or schema at each generation step. Grammar-constrained decoding~\citep{dong2025xgrammar,park2024grammar} suppresses syntactically invalid tokens, while state-tracked constrained decoding~\citep{wang2024fantastic} restricts generation to valid API names and parameters. These methods guarantee well-formed outputs by construction.

\paragraph{Non-token actions.}
An emerging line of work moves beyond the token level entirely by replacing vocabulary tokens with compact latent representations. The BWArea model~\citep{jia2024bwarea} and its successor CoLA~\citep{jia2025controlling} learn a latent action space via an inverse dynamics model; a policy selects latent actions, which a language world model then decodes into tokens. This reduces the action dimensionality but has so far been validated only at moderate model scales.

\subsection{Termination of an Episode}
\label{section:termination}

In classic RL, the termination of an episode is typically determined by the environment: an episode ends when the agent reaches a goal state, fails, or exhausts a fixed step budget.
In LLM training, termination corresponds to the model emitting an end-of-sequence token, at which point the response is complete and a reward can be assigned.
While this appears straightforward, the length of the generated response, and thus the effective episode horizon, is not fixed in advance but emerges from the model's own policy.
This creates a design challenge absent from classic RL: the episode length is both an output of the policy and a factor that influences training dynamics.

Longer episodes give the model more tokens to reason and self-correct, which can improve accuracy on complex tasks.
However, excessively long generations risk verbosity, error accumulation, wasted computation~\citep{su2025between} and memory issues~\citep{zhou2025simple}.
Several recent works have proposed methods to control generation length during RL training.
ThinkPrune~\citep{hou2025thinkprune} imposes a hard token-count cutoff and assigns zero reward to any response that exceeds it; the cutoff is iteratively tightened across training rounds so the model learns to compress its reasoning.
Other approaches apply soft constraints through reward shaping.
LCPO~\citep{aggarwal2025l1} combines correctness rewards with length-based penalties, training the model to respect a target generation budget specified in the prompt.
LEASH~\citep{li2025leash} formulates length control as a constrained optimization problem and uses a Lagrangian dual variable to adaptively adjust penalty strength: the penalty intensifies when generations exceed the target and relaxes when they are shorter.
DAPO~\citep{yu2025dapo} introduces a soft overlong punishment that applies a graduated penalty when the response length exceeds a predefined threshold, with longer overruns receiving proportionally larger penalties.

\subsection{Discount Factor}
\label{section:discount_factor}

The discount factor $\gamma$ determines how future rewards are weighted relative to immediate ones.
In standard RL, the return from time step $t$ is defined as
\[
G_t = \sum_{k=0}^{T-t-1} \gamma^k R_{t+k+1},
\]
where $T$ may be finite or infinite.
For infinite-horizon problems, $\gamma < 1$ is necessary to keep the return bounded; for finite-horizon problems such as LLM generation, the return remains finite even with $\gamma = 1$.

In practice, nearly all LLM training frameworks set $\gamma = 1$~\citep{vonwerra2022trl,sheng2024hybridflow}, and alternative values are rarely explored \citep{ayoub2025learning}.
In many setups, the model receives a single reward $R_T$ at the end of the episode (e.g., a correctness score), with all intermediate rewards equal to zero.
Under this scheme, $G_t = \gamma^{T-t-1} R_T$ for all $t$, so with $\gamma = 1$ the return is simply $R_T$ regardless of response length.
Setting $\gamma < 1$ would instead discount the terminal reward by $\gamma^{T-1}$, making it smaller for longer responses.
This creates an implicit preference for shorter outputs, offering an alternative mechanism for length control that does not require explicit reward shaping.

While $\gamma = 1$ is standard and mathematically sound for finite-horizon LLM training, the potential of $\gamma < 1$ as a lightweight length regularizer remains largely unexplored.
Prior work in traditional RL has studied discounting from multiple angles, including adaptive schedules~\citep{franccois2015discount,fedus2019hyperbolic}, the regularizing effect of $\gamma$~\citep{amit2020discount}, and its theoretical implications~\citep{pitis2019rethinking,tang2021taylor}, but these ideas have yet to be systematically applied to LLM training.

\subsection{Discussion}
\label{section:mdp_discussion}

Across the preceding subsections, a consistent picture emerges: the MDP formulation used in most current LLM training is straightforward, i.e raw token sequences as states, the full vocabulary as actions, $\gamma = 1$, and episode termination at the end-of-sequence token. This default has proven effective in practice~\citep{ouyang2022training,guo2025deepseek}, and its simplicity is itself an advantage since it avoids task-specific engineering.

Meanwhile, each subsection also identifies cases where departing from this default has shown benefits. Compressing the state via summarization~\citep{lu2025scaling} or latent embeddings~\citep{hao2024training} alleviates context-length bottlenecks. Constraining the action space through grammar-guided decoding~\citep{dong2025xgrammar,wang2024fantastic} guarantees structural validity. These results suggest that there is room to incorporate more task structure into the MDP formulation.

A few directions suggested by the current literature may be worth further investigation.

\paragraph{Scaling latent state and action methods.} Latent-state reasoning~\citep{hao2024training,su2025token} and latent-action models~\citep{jia2024bwarea,jia2025controlling} have so far been validated at moderate model scales. It remains to be seen whether their benefits hold as models and datasets grow larger.

\paragraph{Interaction between length-control mechanisms.} Current approaches to generation-length control rely on reward shaping while keeping $\gamma = 1$. Since the discount factor provides a mathematically distinct mechanism for penalizing long episodes, it may be worth studying how reward-based and discount-based length control interact, and whether combining them offers any practical advantage over either alone.

\begin{takeaway}
\textcolor{tabblue}{\textsf{Take-away}} \textbf{The standard MDP formulation is a reasonable default, not a constraint}: the token-level, full-vocabulary, undiscounted setup is sufficient for strong results, but each of its components, state representation, action space, termination, and discounting, admits alternatives that have shown benefits in specific settings.
\end{takeaway}

\section{Exploration}
\label{section:exploration}

\begin{figure}[!htb]
    \centering
    \includegraphics[width=0.9\linewidth]{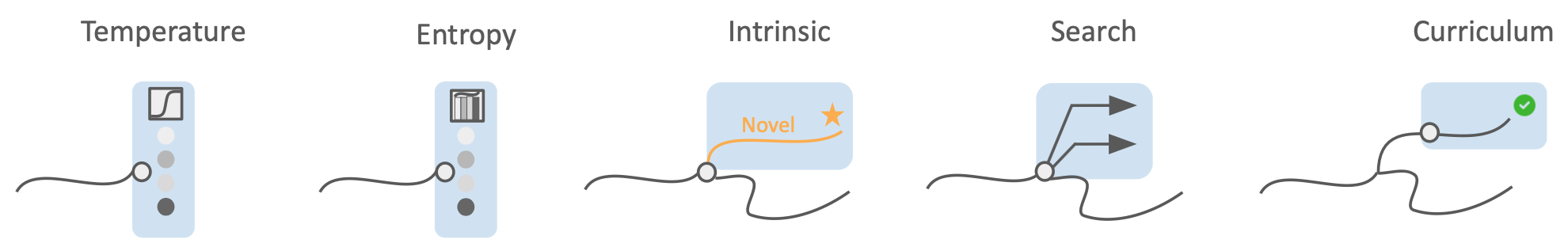}
    \caption{A brief illustration of five exploration strategies in RL-based LLM training, where the shaded region represents the exploration space and the curve represents a reasoning trajectory. Temperature sampling (\cref{section:basic_exploration}) controls the sharpness of the token distribution at each decoding step: lower temperatures concentrate probability mass on likely tokens, while higher temperatures flatten the distribution, inducing broader but undirected exploration. Entropy regularization (\cref{section:entropy}) adds a training-time bonus that discourages the policy from collapsing to deterministic behavior, keeping token distributions closer to uniform across decoding steps. Intrinsic motivation (\cref{section:IM}) provides auxiliary rewards for novel trajectories, steering the agent toward underexplored regions of the space. Tree search (\cref{section:search}) branches systematically at key positions along the trajectory, enabling structured coverage of alternative reasoning paths. Curriculum learning (\cref{section:curriculum_learning}) reduces the effective exploration space by controlling problem difficulty and starting state, thereby concentrating search effort on regions that are learnable at the agent’s current level.}
    \label{fig:exp}
\end{figure}

Exploration is a central challenge in RL for LLM reasoning~\citep{amin2021survey}.
Two properties of reasoning tasks make it particularly difficult.
First, rewards are typically sparse: only the final answer receives a correctness signal, while intermediate reasoning steps go unsupervised.
Second, reasoning problems involve long decision horizons, where each step depends on previous ones and an early mistake can invalidate the entire trajectory.
Since the probability of sampling a correct solution decays exponentially with chain length, na\"ive exploration becomes increasingly ineffective as task complexity grows.
Together, these properties create a setting in which the agent must discover rare, coherent, multi-step trajectories before learning can meaningfully begin.

In policy-based RL algorithms commonly used for LLM training, exploration arises primarily from stochastic sampling over the token vocabulary, modulated by temperature and entropy regularization~\citep{shi2024thorough,ashlag2025state}.
The entropy bonus, incorporated either as an auxiliary loss term~\citep{schulman2017proximal,mnih2016asynchronous} or as part of a maximum-entropy objective~\citep{haarnoja2018soft,ziebart2008maximum}, serves as the primary mechanism for preventing the policy from collapsing to deterministic behavior.

Beyond these basic mechanisms, more structured approaches to exploration have been adapted from classical RL.
Intrinsic motivation methods provide auxiliary reward signals based on novelty or curiosity~\citep{pathak2017curiosity,burda2018exploration}, encouraging the agent to visit underexplored parts of the trajectory space.
Tree search, following the AlphaZero paradigm~\citep{silver2017mastering}, systematically branches and backtracks through partial reasoning paths, enabling the discovery of trajectories that sequential sampling would miss.
Curriculum learning~\citep{narvekar2020curriculum} takes an indirect approach by structuring the sequence of training problems so that the agent encounters progressively more challenging tasks, effectively narrowing the exploration space over time.

Figure \ref{fig:exp} provides an overview of the different components discussed in this section.

\subsection{Temperature Sampling}
\label{section:basic_exploration}

The most basic form of exploration in LLM training arises from stochastic token sampling~\citep{ouyang2022training}.
At each decoding step $t$, the model defines a categorical distribution over the vocabulary $\mathcal{V}$:
\begin{equation}
    \pi_\theta(a_t \mid s_t) = \mathrm{softmax}\!\left(\frac{z_t}{\tau}\right),
    \qquad
    \pi_\theta(a_t = v \mid s_t) = \frac{\exp(z_{t,v}/\tau)}{\sum_{u \in \mathcal{V}} \exp(z_{t,u}/\tau)},
\end{equation}
where $z_t$ denotes the logit vector produced by the model for state $s_t$ and $\tau > 0$ is the temperature.
Rather than selecting the most likely token $a_t^\star = \arg\max_{v \in \mathcal{V}} \pi_\theta(v \mid s_t)$, the model samples
\begin{equation}
    a_t \sim \pi_\theta(\cdot \mid s_t),
\end{equation}
introducing stochasticity into the generation process.
The temperature $\tau$ controls the degree of randomness: lower values sharpen the distribution toward greedy decoding ($\tau \to 0$), while higher values allow the model to explore alternative outputs and encourage creativity or variation. However, temperature-based sampling does not systematically encourage structural or deep exploration, which motivates the development of more principled alternatives.

\subsection{Entropy Regularization}
\label{section:entropy}
While temperature modulates exploration at decoding time, entropy regularization operates directly on the training objective, adding a bonus term to prevent premature convergence to deterministic behavior. The entropy of the policy at state $s_t$ is defined as
\begin{equation}
\mathcal{H}\big(\pi_\theta(\cdot \mid s_t)\big)
= - \sum_{v \in \mathcal{V}}
\pi_\theta(v \mid s_t) \,
\log \pi_\theta(v \mid s_t),
\label{eq
:entropy}
\end{equation}
weighted by a tunable coefficient~\citep{vonwerra2022trl,sheng2024hybridflow,zhang2024framework}. Too small a weight fails to prevent entropy collapse; too large introduces excessive randomness. A key challenge is therefore not just how much entropy regularization to apply, but where.
\paragraph{Entropy is not uniform across reasoning trajectories.}
Planning and transitional tokens exhibit higher entropy than execution tokens, and elevated entropy tends to signal underexplored solution paths~\citep{cheng2025reasoning}. This means the exploration signal is concentrated at a small subset of positions, making uniform entropy regularization across all tokens inefficient. \citet{cheng2025reasoning} exploit this by shaping the advantage function with an entropy-based term at high-entropy positions rather than adding a uniform bonus to the loss. \citet{cui2025entropy} provide a complementary theoretical account: entropy dynamics are dominated by high-covariance tokens, and naively regularizing all tokens can therefore be ineffective; they propose penalizing updates specifically to these tokens for more targeted control. Together, these results suggest that the unit of entropy regularization should be the token, not the sequence.
\paragraph{Controlling entropy in practice.}
Given that entropy is both non-uniform and difficult to set statically, several works address the practical challenges of applying it during training. \citet{he2025skywork} propose an adaptive scheme that adjusts the entropy coefficient to stabilize training around a target level. \citet{shen2025entropy} reduce the computational cost of full-vocabulary entropy by evaluating a clamped term over a reduced, input-dependent token set. PREPO~\citep{sun2025efficient} takes a sequence-level view, aggregating token-level entropy into a score that reweights gradient updates toward more uncertain rollouts.
While entropy regularization is well understood in classical RL~\citep{haarnoja2017reinforcement,ahmed2019understanding,haarnoja2018soft}, its role in LLM training remains less clear. The finding that strong performance is achievable both with~\citep{agarwal2025unreasonable} and without~\citep{hu2025open} explicit entropy bonuses suggests its importance is highly context-dependent, and several recent works call for further investigation into how entropy dynamics interact with learning~\citep{agarwal2025unreasonable,cui2025entropy}.

\subsection{Intrinsic Motivation}
\label{section:IM}

\begin{table}[!htb]
    \centering
    \begin{tabular}{l|l|l}
        \toprule
        \textbf{Method} & \textbf{Novelty Granularity} & \textbf{RL Counterpart} \\
        \midrule
    CDE~\citep{dai2025cde} & Token & RND~\citep{burda2018exploration} \\
        CD-RLHF~\citep{sun2025curiosity} & Token & ICM~\citep{pathak2017curiosity} \\
        i-MENTOR~\citep{gao2025navigate} & Trajectory &  RND \\
        MERCI~\citep{zhang2025count} & Trajectory & Pseudo-count~\citep{lobel2023flipping} \\
        DACE~\citep{li2025know} & Trajectory & Uncertainty \\
        INTUITOR~\citep{zhao2025learning} & Trajectory & Uncertainty \\
        \bottomrule
    \end{tabular}
    \caption{Intrinsic motivation methods in RL-based LLM training, grouped by the granularity at which novelty is operated.}
    \label{tab:im_def}
\end{table}

Inspired by how humans explore through intrinsic motivation~\citep{aubret2019survey,chentanez2004intrinsically}, classical RL rewards novelty via prediction errors~\citep{pathak2017curiosity}, information gain~\citep{houthooft2016vime}, or model disagreement~\citep{sekar2020planning}. Adapting these ideas to LLM training raises a non-trivial question: what counts as novelty for a language model? The methods in~\cref{tab:im_def} reflect three implicit answers, differing in the granularity at which novelty is measured.

\textbf{Token-level} methods treat low-probability outputs as novel. CD-RLHF~\citep{sun2025curiosity} applies an ICM-style forward dynamics error selectively to tokens outside the top-$k$ most likely actions. CDE~\citep{dai2025cde} combines response perplexity with value variance from a multi-head critic. Both are computationally cheap but conflate lexical surprise with meaningful exploration: a model can inflate its novelty score by generating incoherent outputs.

\textbf{Trajectory-level} methods aggregate signals over full reasoning sequences. i-MENTOR~\citep{gao2025navigate} computes prediction-error-based rewards over full reasoning sequences, applied selectively to incorrect trajectories. MERCI~\citep{zhang2025count} estimates pseudo-counts over trajectories via a Coin Flipping Network~\citep{lobel2023flipping}, directly penalising repetition. These signals are more semantically meaningful than token-level alternatives, but whether pseudo-counts and prediction errors reliably capture meaningful novelty over reasoning trajectories remains an open question. A different approach is to replace novelty with confidence: DACE~\citep{li2025know} aggregates token-level certainty into a response-level scalar and uses it adaptively, penalising high certainty on difficult problems and rewarding it on easy ones. INTUITOR~\citep{zhao2025learning} goes further, using self-certainty as the sole training signal with no extrinsic reward. This shift suggests that for pretrained LLMs, where novelty is hard to define, confidence may be a more natural proxy for whether a model needs to explore further.

Several challenges familiar from classical RL carry over to this setting.
Balancing intrinsic and extrinsic reward magnitudes requires careful tuning~\citep{sun2025curiosity}, and most current methods rely on fixed hyperparameters that may not generalize across tasks or training stages; adaptive scheduling~\citep{lee2022adaptive} may help.
Na\"ively combining intrinsic rewards with the task objective can also distort the learning signal.
In classical RL, decoupling exploration and exploitation through separate policies has been shown to stabilize training~\citep{schafer2021decoupled}, and similar ideas may be applicable in the LLM setting.

\paragraph{Benchmarking and combining intrinsic motivation methods.}
The five methods in \cref{tab:im_def} each adapt a different classical signal (prediction error, pseudo-counts, uncertainty) but no existing work evaluates them on a shared benchmark.
A systematic comparison would clarify which signals are complementary and which are redundant.
If they do capture different aspects of novelty, combining multiple intrinsic objectives within a single training run~\citep{yuan2025deep} would be worth investigating.

\subsection{Search}
\label{section:search}

While intrinsic motivation encourages exploration through reward shaping, tree search takes a more structural approach: explicitly branching and backtracking through partial reasoning paths to diversify the trajectories seen during RL training, drawing on the AlphaZero paradigm~\citep{silver2017mastering} where MCTS acts as a policy-improvement operator. Search has proven effective at improving reasoning in related settings, both at inference time and for generating SFT and preference learning data~\citep{zhang2024rest, chen2024alphamath, xie2024monte}, but its integration into the RL training loop itself remains a largely open research area.

The few existing works that do so share a common principle: rather than resampling full trajectories independently, they branch selectively at high-uncertainty tokens, concentrating the exploration budget at genuine divergence points. TreeRL~\citep{hou2025treerl} derives on-policy process supervision signals from the resulting tree, eliminating the need for a separately trained process reward model. TreePO~\citep{li2025treepo} reuses the KV cache across branches to reduce inference cost by up to $43\%$, and derives segment-level advantage estimates that improve credit assignment over standard GRPO. BFS-PO~\citep{parascandolo2026bfs} conditions each new branch on the shortest correct solution found so far, steering the policy toward more concise reasoning. All three demonstrate that branching at divergence points is both more efficient and more informative than uniform resampling.

Given the strong results of search in inference-time scaling and data synthesis, its integration into RL training presents many promising research directions. 

\subsection{Curriculum Learning}
\label{section:curriculum_learning}

\begin{table}[!htb]
    \centering
    \small
    \begin{tabular}{l|l|l}
        \toprule
        \textbf{Category} & \textbf{Method} & \textbf{Mechanism} \\
        \midrule
        Data Sampling & E2H~\citep{parashar2025curriculum} & Easy-to-hard scheduling based on success rate \\
        - & Re-Schedule~\citep{wang2025scheduling} & Simple-to-complex scheduling based on reasoning tree structure \\
        - & DARS~\citep{yang2025depth} & Allocate more compute to harder (low-accuracy) problems \\
        Starting State & R$^3$~\citep{xi2024training} & Start near solution, progressively move backward \\
        - & MARGE~\citep{gaomarge} & Explore from previously successful intermediate states \\
        - & FR3E~\citep{zheng2025first} & Targeted rollouts from high-entropy (uncertain) positions \\
        Reward Signal & DELTA-Code~\citep{sun2025delta} & Dense (per-test pass rate) then sparse (binary correctness) \\
        - & SRL~\citep{deng2025supervised} & Dense (action--expert similarity) then sparse (outcome correctness) \\
        \bottomrule
    \end{tabular}
    \caption{Curriculum learning methods in RL-based LLM training.}
    \label{tab:curriculum}
\end{table}

Curriculum learning~\citep{narvekar2020curriculum} can be viewed as an indirect form of exploration: rather than modifying action selection to better explore, it shapes the distribution of training experiences so that the agent encounters problems of appropriate difficulty at each stage, simplifying the exploration.
In classical RL, curricula have been designed along three dimensions: (1)~data sampling, scheduling tasks of varying difficulty across training~\citep{fang2019curriculum,portelas2020automatic}; (2)~starting state, initializing the agent closer to goal states to simplify early exploration~\citep{uchendu2023jump,chane2021goal}; and (3)~reward signal, beginning with dense rewards and transitioning to sparse ones~\citep{park2025sparse}.
Recent work on RL-based LLM training has adopted analogous strategies along all three dimensions, as summarized in~\cref{tab:curriculum}.

A key complementary idea across these methods is the use of teacher models as domain-specific experts to stabilize and enrich learning signals. Most current training pipelines such as  Nemotron-3 \cite{nvidia_nemotron3_ultra_2026} train specialized teacher models, each optimized through its own domain-specific training pipeline. Rather than relying solely on sparse environment rewards, teacher models can then provide structured supervision that reflects higher-quality reasoning in specialized domains. This enables more effective knowledge transfer via distillation.  The student model generates rollouts across all domains and receives dense reward signals from the corresponding teacher models. The co-evolution between student and teachers enables continuous capability improvement and progressively stronger specialization across domains.

\paragraph{Data sampling.}
E2H~\citep{parashar2025curriculum} and Re-Schedule~\citep{wang2025scheduling} both implement easy-to-hard scheduling but differ in their difficulty metrics: E2H ranks problems by success rate, while Re-Schedule uses a reasoning complexity score derived from reasoning trees.
DARS~\citep{yang2025depth} takes a complementary approach, first estimating problem difficulty through rollouts and then dynamically allocating additional compute to harder problems with low accuracy.

\paragraph{Starting state.}
R$^3$~\citep{xi2024training} introduces a reverse curriculum that initializes the reasoning process near a demonstration's end and progressively moves the starting point backward, so the model first learns from states close to the reward before tackling more distant starting points.
MARGE~\citep{gaomarge} expands exploration from intermediate states where the model has previously succeeded.
FR3E~\citep{zheng2025first}, inspired by Go-Explore~\citep{ecoffet2021first}, a strong structural exploration technique, identifies positions along trajectories where token-level entropy is highest and performs targeted rollouts from these states to discover alternative solution paths.

\paragraph{Reward signal.}
Both DELTA-Code~\citep{sun2025delta} and SRL~\citep{deng2025supervised} implement dense-to-sparse reward curricula.
DELTA-Code begins with per-test pass rates before transitioning to binary correctness rewards.
SRL starts with dense step-level rewards based on similarity to expert demonstrations, then transitions to sparse outcome-based rewards. 

\subsection{Discussion}
\label{section:exploration_discussion}

The five exploration strategies described above, including temperature sampling, entropy regularization, intrinsic motivation, tree search, and curriculum learning, all encourage exploration, but most operate primarily at the token or local decision level rather than directly at the level of abstract reasoning strategies. Temperature sampling and entropy regularization promote diversity in token choices at each step, and intrinsic motivation rewards novel trajectories, yet a model can still generate many variations of a solution that follow the same high-level strategy. Tree search provides more structured branching, but many branches may explore superficially different token sequences rather than fundamentally distinct approaches. Curriculum learning shapes the exploration space by adjusting problem difficulty, but it does not guarantee strategic diversity within that space. Encouraging trajectory-level or strategy-level variation, meaning distinct paths through the reasoning space rather than only token-level diversity, remains a key opportunity for improving exploration in RL-based LLM training.

\begin{takeaway}
\textcolor{tabblue}{\textsf{Take-away}} \textbf{Exploration in LLM training remains fragmented}: existing methods address complementary dimensions of the exploration challenge, such as token-level randomness, novelty signals, structured search and problem selection. Yet, they are usually used separately, and there is limited empirical evidence comparing their performance or assessing their potential synergies and trade-offs in integrated settings. In addition, many approaches still operate primarily at the token or local decision level, and do not explicitly encourage strategy- or trajectory-level diversity, meaning that many superficially different outputs may follow the same underlying reasoning path.
\end{takeaway}


\section{Learning}
\label{section:learning}


\begin{figure}[!htb]
    \centering
    \includegraphics[width=0.6\linewidth]{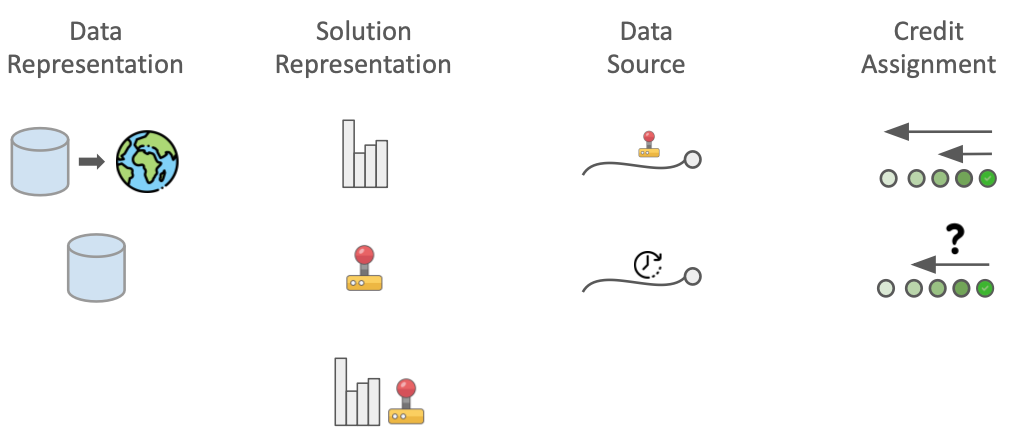}
    \caption{An illustration of the four key design dimensions in RL-based LLM training. Data representation (\cref{section:model_free_based}) distinguishes model-free methods, which update the policy directly from collected trajectories, from model-based methods, which first learn a world model. Solution representation (\cref{section:value_policy}) categorizes algorithms by whether they learn a value function, a policy, or both (actor-critic). Data source (\cref{section:on_off_policy}) contrasts on-policy methods, which train on freshly sampled rollouts, with off-policy methods, which reuse trajectories from earlier or different policies. Credit assignment (\cref{section:credit_assignment}) addresses how the outcome of a trajectory is attributed back to individual tokens, and what value should be used for backpropagation.}
    \label{fig:learning}
\end{figure}

The learning component involves several key design decisions. Model-free methods train the policy or value function directly on observed trajectories, while model-based methods first learn a dynamics model and then generate additional training signals. In model-free, policies can be represented alone (as in REINFORCE~\citep{williams1992simple}), value functions alone (as in Q-learning~\citep{watkins1992q}), or both together, as in actor-critic approaches such as PPO~\citep{schulman2017proximal}. The source of training data also matters: on-policy methods use trajectories generated by the current policy $\pi_\theta$, whereas off-policy methods can reuse trajectories collected by previous or alternative policies. Learning often occurs in parallel with trajectory collection, allowing the agent to improve continuously as more experience is gathered. Finally, credit assignment specifies how the outcome of a trajectory is attributed to individual actions $a_t$ along the trajectory.


\subsection{Data Representation: Model-Free or Model-Based Methods}
\label{section:model_free_based}

A central distinction in RL is whether collected data is used to learn a dynamics model or to update the policy directly.
Model-based methods such as Dyna~\citep{sutton1991dyna}, Dreamer~\citep{hafner2025mastering}, and TD-MPC~\citep{hansen2023td} take the former route: they learn a transition model from experience and optimize the policy using simulated trajectories.
Model-free methods such as DQN~\citep{mnih2013playing}, REINFORCE~\citep{williams1992simple}, and PPO~\citep{schulman2017proximal} take the latter, updating $\pi_\theta$ or the value function directly from collected experience.
Nearly all RL fine-tuning of LLMs follows the model-free approach: algorithms such as PPO and GRPO~\citep{shao2024deepseekmath} update $\pi_\theta$ directly from sampled rollouts without learning a separate dynamics model during the RL phase.
An interesting perspective is that the pretrained LLM can be viewed as an implicit language dynamics model, since it was trained to predict $\pi_\theta(a_t \mid s_t)$ over large text corpora.
Under this interpretation, if we regard the base policy as a learned transition model of language, then most current RL fine-tuning of LLMs could be seen as a form of model-based RL, where pretraining builds the world model and RL optimizes a policy on top of it. This offers another perspective of looking at RL fine-tuning of LLMs.

\subsection{Solution Representation: Value Function, Policy, or Both}
\label{section:value_policy}

\begin{figure}[!htb]
\centering
\begin{tikzpicture}[
    font=\small,
    every node/.style={align=center},
    category/.style={
        rounded corners=8pt,
        draw=none,
        minimum width=3.6cm,
        minimum height=0.75cm,
        font=\small\bfseries\color{white},
    },
    method/.style={
        rounded corners=4pt,
        draw=gray!40,
        fill=white,
        minimum width=3.6cm,
        minimum height=0.62cm,
        font=\small,
        text width=3.4cm,
    },
]

\definecolor{muteblue}{HTML}{378ADD}
\definecolor{warmgray}{HTML}{5F5E5A}
\definecolor{rose}{HTML}{72243E}

\def\vsep{0.18cm}

\node[category, fill=muteblue] (vb) {Value-based};

\node[method, below=0.35cm of vb]    (qsft)  {Q-SFT \citep{hong2024q}};
\node[method, below=\vsep of qsft]   (tbrm)  {TBRM \citep{yuan2025trajectory}};
\node[method, below=\vsep of tbrm]   (shiq)  {ShiQ \citep{clavier2025shiq}};
\node[method, below=\vsep of shiq]   (qsh)   {Q$^\sharp$ \citep{zhou2025q}};
\node[method, below=\vsep of qsh]    (tq)    {Transfer Q$^*$ \citep{chakraborty2024transfer}};
\node[method, below=\vsep of tq]     (ilql)  {ILQL \citep{snell2022offline}};

\node[category, fill=warmgray, right=1.8cm of vb] (pb) {Policy-based};

\node[method, below=0.35cm of pb]   (grpo)    {GRPO \citep{shao2024deepseekmath}};
\node[method, below=\vsep of grpo]  (remax)   {ReMax \citep{li2023remax}};
\node[method, below=\vsep of remax] (rloo)    {RLOO \citep{ahmadian2024back}};
\node[method, below=\vsep of rloo]  (variant) {Variants of GRPO:\\
                                                GSPO~\citep{zheng2025group}\\
                                                GFPO~\citep{shrivastava2025sample}, 
                                                GDPO~\citep{liu2026gdpo}, etc.};

\node[category, fill=rose, right=1.8cm of pb] (ac) {Actor-critic};

\node[method, below=0.35cm of ac]    (ppo)    {PPO \citep{schulman2017proximal}};
\node[method, below=\vsep of ppo]    (asyppo) {AsyPPO \citep{liu2025asymmetric}};
\node[method, below=\vsep of asyppo] (tppo)   {T-PPO \citep{fan2025truncated}};
\node[method, below=\vsep of tppo]   (vcppo)  {VC-PPO \citep{yuan2025s}};
\node[method, below=\vsep of vcppo]  (vapo)   {VAPO \citep{yue2025vapo}};
\node[method, below=\vsep of vapo]   (ppom)   {PPO-max \citep{zheng2023secrets}};
\node[method, below=\vsep of ppom]   (vrpo)   {VRPO \citep{zhu2025vrpo}};
\node[method, below=\vsep of vrpo]   (dvpo)   {DVPO \citep{zhu2025dvpo}};

\begin{scope}[on background layer]
    \node[rounded corners=10pt, fill=muteblue!8, draw=muteblue!30,
          fit=(vb)(ilql), inner sep=10pt] {};
    \node[rounded corners=10pt, fill=warmgray!8, draw=warmgray!30,
          fit=(pb)(variant), inner sep=10pt] {};
    \node[rounded corners=10pt, fill=rose!8, draw=rose!30,
          fit=(ac)(dvpo), inner sep=10pt] {};
\end{scope}

\end{tikzpicture}
\caption{Categorization of RL methods for LLM training by solution representation.}
\label{fig:rl_taxonomy}
\end{figure}
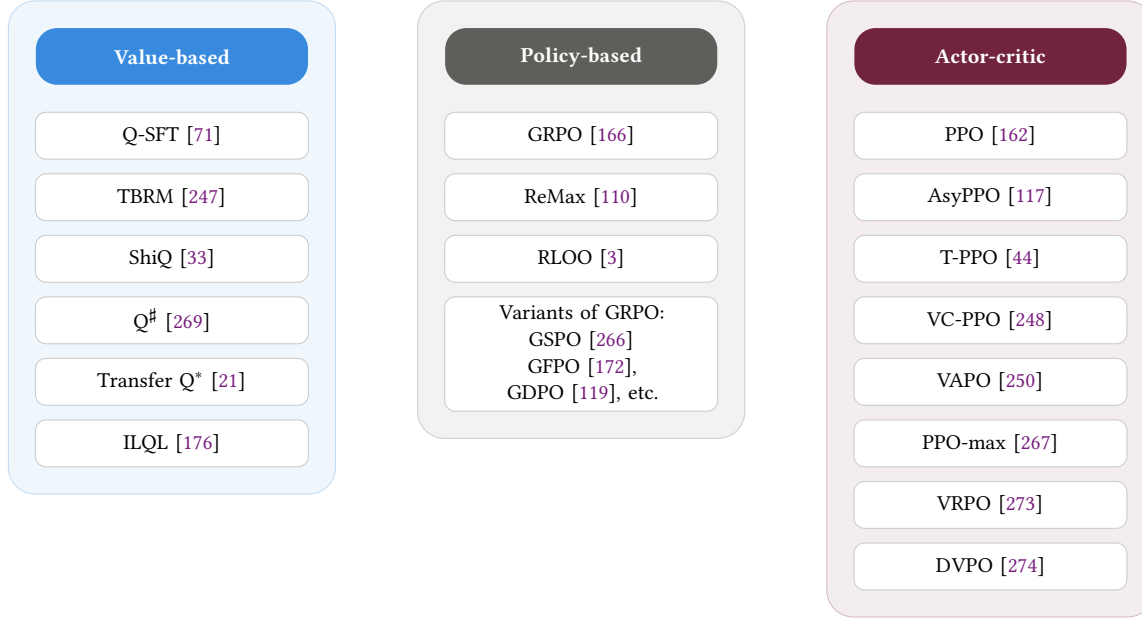

RL algorithms can be categorized by whether they learn a value function, a policy, or both.
Value-based methods estimate the expected return from a state $s_t$ or state-action pair $(s_t, a_t)$ and derive a policy implicitly, typically by choosing actions greedily with respect to the estimated values.
Classic examples include DQN~\citep{mnih2013playing}, Rainbow~\citep{hessel2018rainbow}, and PQN~\citep{gallici2024simplifying,kooi2025hadamax}.
Policy-based methods directly parameterize and optimize the policy $\pi_\theta$ via the policy gradient theorem, without requiring an explicit value function; REINFORCE~\citep{williams1992simple} and GRPO~\citep{shao2024deepseekmath} are representative examples.
Actor-critic methods combine both: the actor maintains a parameterized policy and the critic learns a value function, used either for variance reduction in the policy gradient (e.g., PPO~\citep{schulman2017proximal}) or for directly guiding the actor's updates (e.g., SAC~\citep{haarnoja2018soft}). 

Since LLMs are trained as next-token predictors, they naturally define an autoregressive policy $\pi_\theta(a_t \mid s_t)$ over the vocabulary (see~\cref{section:llm_concepts}).
This makes policy-based and actor-critic methods the most straightforward choice, and indeed they dominate current practice.
Value-based methods require additional design effort to integrate with the LLM architecture, but recent work has shown that they can offer complementary benefits such as improved sample efficiency and off-policy learning.
We review each category below.

\paragraph{Value-based methods.}
A canonical value-based algorithm is Q-learning~\citep{watkins1992q}, which iteratively updates value estimates using a sample-based approximation of the Bellman optimality equation:
\begin{equation}
Q(s_t, a_t) \leftarrow Q(s_t, a_t) + \alpha \left[ R_{t+1} + \gamma \max_{a'} Q(s_{t+1}, a') - Q(s_t, a_t) \right],
\end{equation}
where $\alpha \in (0, 1]$ is the learning rate.
Applying value-based methods to LLMs is not straightforward, since the model is parameterized as a policy rather than a value function.
Recent work addresses this in two ways.

The first approach trains the LLM itself with value-learning objectives, exploiting the observation that under KL-regularized RL, the difference between current and reference logits is proportional to Q-values.
Q-SFT~\citep{hong2024q} casts Q-learning as modified supervised fine-tuning where token probabilities directly encode Q-values, enabling a seamless transition from pre-training to value learning.
TBRM~\citep{yuan2025trajectory} extends Q-learning to the trajectory level, optimizing a single TD error over the entire response rather than per-token errors.
ShiQ~\citep{clavier2025shiq} uses $n$-step returns for improved credit assignment, with provable convergence guarantees.

The second approach trains a dedicated value network while keeping the LLM weights frozen.
Q$^\sharp$~\citep{zhou2025q} learns the optimal Q-function via distributional RL and steers the frozen policy by adding Q-values to reference logits at inference time.
Transfer~Q$^*$~\citep{chakraborty2024transfer} estimates Q-values implicitly through a pre-aligned baseline model without any fine-tuning.
ILQL~\citep{snell2022offline} learns Q-functions offline with conservative Bellman backups and uses them to adjust logits toward reward-maximizing tokens at inference.

Although integrating value-based methods into LLM training requires additional design effort, recent work has demonstrated advantages over policy-based approaches, including improved sample efficiency~\citep{clavier2025shiq,yuan2025trajectory,snell2022offline}, greater training stability~\citep{zhou2025q,snell2022offline,hong2024q}, and provable convergence guarantees under mild assumptions~\citep{clavier2025shiq,zhou2025q}.

\paragraph{Policy-based methods.}
Policy gradient methods optimize the expected return by ascending the gradient~\citep{sutton2000policy,schulman2015high,weng2018PG}:
\begin{equation}
\label{eq:policy_gradient}
g = \mathbb{E}\left[ \sum_{t=0}^{T} Q^{\pi}(s_t, a_t) \nabla_\theta \log \pi_\theta(a_t \mid s_t) \right],
\end{equation}
where $Q^{\pi}(s_t, a_t)$ denotes the expected return for a given state-action pair and serves as a weighting factor that determines how the policy parameters should be updated.
Actions yielding higher returns get increased probability, while those yielding lower returns get decreased probability.

In practice, $Q^{\pi}$ is often replaced by the advantage function $A^{\pi}(s_t, a_t) = Q^{\pi}(s_t, a_t) - V^{\pi}(s_t)$, which measures how much better a specific action is relative to the average under the current policy.
Using the advantage reduces variance while preserving an unbiased gradient~\citep{schulman2015high}.
This subsection focuses on methods that estimate $Q^{\pi}$ or $A^{\pi}$ solely from sampled rollouts, without learning a separate value network.

ReMax~\citep{li2023remax} is an early advocate of critic-free RL for LLMs, building on REINFORCE with a variance reduction baseline estimated by greedily sampling a single response and using its reward as the baseline value.
By removing the value model entirely, ReMax halves GPU memory usage and doubles training throughput compared to PPO, while matching its task performance.
\citet{ahmadian2024back} further demonstrate that a critic network is unnecessary for LLM alignment and employ REINFORCE with Leave-One-Out (RLOO)~\citep{kool2019buy}, which samples multiple completions per prompt and uses the mean reward of the remaining $k{-}1$ completions as the baseline for each sample.
This yields a low-variance, unbiased gradient estimate that consistently outperforms PPO without requiring a learned value function.

GRPO~\citep{shao2024deepseekmath} takes a different approach to critic-free advantage estimation by sampling multiple responses per prompt and computing advantages relative to the group mean, again eliminating the need for a learned value function.
This design has inspired a rich family of variants that refine different components of the algorithm, such as advantage normalization, clipping strategy, and loss granularity~\citep{zheng2025group,shrivastava2025sample,zhao2025geometric,hu2025reinforce++}.
Since these methods target specific aspects of GRPO, we defer their discussion to~\cref{section:credit_assignment}.

\paragraph{Actor-critic methods.}
Actor-critic methods learn both a policy (the actor) and a value function (the critic).
The critic can serve two distinct purposes:
(1)~variance reduction, where the critic provides a baseline for the policy gradient, yielding more accurate advantage estimates in~\cref{eq:policy_gradient}, methods such as A2C~\citep{mnih2016asynchronous}, TRPO~\citep{schulman2015trust}, and PPO~\citep{schulman2017proximal} follow this approach, and are sometimes termed critic-based methods~\citep{zhang2025survey} in contrast to the critic-free approaches described above (policy-based methods);
and (2)~direct policy learning, where the actor is trained to maximize the critic's output by backpropagating gradients through the Q-function, DDPG~\citep{lillicrap2015continuous}, TD3~\citep{fujimoto2018addressing}, and SAC~\citep{haarnoja2018soft} follow this approach.
In the LLM setting, existing actor-critic methods exclusively use the critic for variance reduction in the PPO style~\citep{ouyang2022training,chen2024improving,hu2024improving}, while SAC-style direct policy learning through the critic remains unexplored.
We therefore focus on how recent work has improved PPO-style training for LLMs along three axes: computational overhead, value estimation bias, and training stability.

A first line of work addresses the computational overhead of maintaining a separate critic.
AsyPPO~\citep{liu2025asymmetric} replaces the same-size critic with an ensemble of lightweight mini-critics trained on disjoint prompt shards, and leverages inter-critic disagreement to mask uninformative advantages, reducing system overhead by over 20\%.
T-PPO~\citep{fan2025truncated} takes an orthogonal approach by applying Extended GAE to estimate advantages from partial rollouts, enabling policy updates before full generation completes and achieving up to 2.5$\times$ training speedup.

A second line of work addresses value estimation bias, which becomes especially acute in long chain-of-thought tasks.
VC-PPO~\citep{yuan2025s} identifies two sources: initialization bias from the common practice of deriving the value model from the reward model, and reward signal decay where GAE's exponential discounting prevents reward information from reaching earlier tokens.
It addresses these by pretraining the critic under a fixed policy and decoupling the GAE computation so that the policy and value networks use different discount parameters.
VAPO~\citep{yue2025vapo} extends this with length-adaptive GAE that adjusts the discount factor based on sequence length, and incorporates clip-higher and dynamic group sampling to address reward sparsity.

A third line of work focuses on training stability and robustness.
PPO-max~\citep{zheng2023secrets} systematically evaluates implementation choices across PPO components and identifies the most effective configuration for each.
VRPO~\citep{zhu2025vrpo} combines an auxiliary loss with a variational information bottleneck to improve the critic's robustness under noisy reward signals.
DVPO~\citep{zhu2025dvpo} observes that both mean-based methods (PPO, GRPO) and worst-case methods can be overly conservative under noisy rewards, and instead learns token-level value distributions with asymmetric risk regularization to dampen noisy deviations while preserving exploratory diversity.
A summary of these methods and the challenges they address is provided in~\cref{tab:ac}.

\begin{table}[!htb]
    \centering
    \begin{tabular}{l|l}
    \toprule
        \textbf{Method} & \textbf{Challenge Addressed}\\
        \midrule
        PPO~\citep{schulman2017proximal} & -- \\
        AsyPPO~\citep{liu2025asymmetric} & Computational and memory overhead \\
        T-PPO~\citep{fan2025truncated} & Training inefficiency on long sequences \\
        VC-PPO~\citep{yuan2025s} & Value initialization bias, reward signal decay \\
        VAPO~\citep{yue2025vapo} & Value bias, heterogeneous lengths, reward sparsity \\
        PPO-max~\citep{zheng2023secrets} & Training stability \\
        VRPO~\citep{zhu2025vrpo} & Noisy reward signals \\
        DVPO~\citep{zhu2025dvpo} & Conservative value estimation under noise \\
        \bottomrule
    \end{tabular}
    \caption{Actor-critic methods improving upon PPO for LLM training.}
    \label{tab:ac}
\end{table}

\paragraph{Discussion.}
Across the preceding paragraphs, a clear divergence from conventional RL emerges: whereas actor--critic and value-based methods dominate in classical settings (precisely because pure policy gradients suffer from high variance), critic-free policy-gradient methods such as GRPO~\citep{shao2024deepseekmath} and its variants~\citep{zheng2025group,yu2025dapo} have become the default in LLM training. This reversal can be traced to properties specific to the LLM setting.

First, the LLM environment is fully resettable and supports batch sampling: multiple completions can be sampled from any prompt without costly external interactions. This enables low-variance advantage estimation through large-scale sampling, effectively substituting compute for a learned critic. Second, eliminating the critic substantially reduces memory overhead, since in PPO-style LLM training the critic is typically a full-sized language model that doubles memory requirements~\citep{huang2024n+}. Together, these two factors explain why critic-free methods match or exceed PPO performance in practice while being considerably cheaper to run.

However, actor--critic methods can remain valuable when the setting demands it. For long chain-of-thought tasks, where reward sparsity and signal decay make group-based advantage estimation less reliable, recent PPO variants such as VC-PPO~\citep{yuan2025s} and VAPO~\citep{yue2025vapo} demonstrate clear benefits from a learned value function. Value-based methods offer a different set of advantages (sample efficiency and off-policy learning), though they currently require additional design effort to integrate with the autoregressive LLM architecture.

A notable gap in the current literature is the absence of SAC-style actor--critic methods, where the actor maximizes the critic's Q-estimates via backpropagation rather than using the critic solely for variance reduction. This gap has a concrete explanation: backpropagating through a large critic network is prohibitively expensive, and existing SAC extensions to discrete actions~\citep{christodoulou2019soft} assume small action spaces that do not scale to LLM vocabularies (32K--150K tokens).

Several natural directions for future work follow from the current literature.

\paragraph{Bridging pre-training and value learning.}
LLMs are pre-trained to output a distribution over the vocabulary, but value-based RL requires scalar value estimates for states or state-action pairs. Current methods bridge this gap either by reinterpreting token probabilities as Q-values~\citep{hong2024q} or by attaching a separate value head to the frozen model~\citep{zhou2025q}, both of which introduce design overhead absent from policy-based approaches. Finding a way to incorporate value-aware auxiliary objectives during pre-training could potentially reduce this mismatch and make value-based RL a more natural downstream choice.

\paragraph{Enabling SAC-style training.}
Reducing critic size while maintaining estimation accuracy, through distillation or lightweight critic heads~\citep{liu2025asymmetric}, could make gradient-based actor--critic updates feasible at LLM scale, opening a class of algorithms that is currently inaccessible~\citep{haarnoja2018soft,christodoulou2019soft}.

\paragraph{Characterizing method trade-offs.}
The choice between pure policy-based~\citep{shao2024deepseekmath,ahmadian2024back}, actor--critic~\citep{schulman2017proximal,yuan2025s,yue2025vapo}, and value-based methods~\citep{hong2024q,zhou2025q,clavier2025shiq} currently relies on informal heuristics. Systematic comparisons across task types (short-answer vs.\ long-chain reasoning), reward structures (dense vs.\ sparse), and model scales would help practitioners decide when the overhead of a critic or value network is warranted.

\begin{takeaway}
\textcolor{tabblue}{\textsf{Take-away}} \textbf{Critic-free policy gradients are the practical default, not the only option}: the flexibility of the LLM sampling and the memory cost of a full-sized critic explain why pure policy-based methods like GRPO dominate current practice. Actor-critic and value-based methods have been explored as alternatives, though only in a limited number of works, and their practical role in LLM training remains an open direction.
\end{takeaway}

\subsection{Data Source: On-Policy or Off-Policy?}
\label{section:on_off_policy}

In RL, algorithms are categorized as on-policy or off-policy based on the relationship between the data-generating policy and the policy being improved. On-policy methods, such as PPO~\citep{schulman2017proximal}, TRPO~\citep{schulman2015trust}, and A3C~\citep{mnih2016asynchronous}, learn exclusively from transitions generated by the current policy, discarding data after each update. Off-policy methods, such as DQN~\citep{mnih2013playing}, SAC~\citep{haarnoja2018soft}, and TD3~\citep{fujimoto2018addressing}, can learn from data generated by a different behavior policy, enabling replay of past experience.

When the behavior and target policies differ, the resulting distribution mismatch must be accounted for. Policy-gradient methods typically correct for this via importance sampling (IS), reweighting transitions by the probability ratio between target and behavior policies. It’s good because importance sampling makes updates unbiased by correcting for the mismatch between behavior and target policies, but bad because those probability ratios can have high variance, leading to unstable or inefficient learning. Value-based methods such as Q-learning sidestep explicit IS altogether: the Bellman optimality operator ($\max_a Q(s,a)$) directly estimates the value of the greedy policy regardless of how the data was collected.

The classic \textit{Cliff Walking} task~\citep{sutton1998reinforcement} illustrates the practical consequences of this distinction. The shortest path to the goal runs along a cliff edge with a large penalty for falling. Q-learning (off-policy) converges to this optimal but risky path, because its updates assume a perfectly greedy target policy and ignore the exploratory actions that may cause the agent to fall. SARSA (on-policy), by contrast, converges to a longer but safer path away from the cliff, because it incorporates the agent's actual exploratory behavior into its value estimates and thus accounts for the risk of missteps.

More broadly, on-policy methods produce value estimates that reflect the current policy's behavior, including its exploration noise, while off-policy methods can in principle learn the optimal policy independently of how data is collected. This independence enables greater sample efficiency through experience replay, but it also gives rise to the deadly triad~\citep{van2018deep}: when off-policy updates, function approximation, and bootstrapping are combined, value estimates can diverge and destabilize training.

\paragraph{On-policy.}
Variants of REINFORCE~\citep{ahmadian2024back,hu2025reinforce++}, PPO~\citep{ouyang2022training,hu2025open,yuan2025s}, and GRPO~\citep{shao2024deepseekmath,zheng2025group} are on-policy methods that optimize the same policy used to generate the training data. Although PPO and GRPO employ importance sampling (IS), they remain on-policy because IS is used to enable multiple optimization epochs on the same batch of trajectories, correcting for the minor distribution shift that accumulates within a single iteration rather than learning from an entirely different behavior policy. Most on-policy methods for LLM training are GRPO variants that identify and address specific limitations, such as improving advantage estimation~\citep{simoni2025gtpo,liu2026gdpo} and achieving more accurate importance sampling~\citep{zheng2025group,chu2025gpg}; we discuss these methods in depth in~\cref{section:credit_assignment}.

\paragraph{Off-policy.}
Off-policy methods for LLM training can be organized along three motivations: leveraging external demonstrations, stabilizing optimization with stale data, and improving theoretical foundations.

A first line of work leverages demonstrations from stronger models to accelerate learning. LUFFY~\citep{yan2025learning} mixes off-policy demonstrations with on-policy rollouts during advantage computation, using regularized importance sampling to prevent superficial imitation and entropy collapse. CHORD~\citep{zhang2025policy} takes a complementary approach, reframing SFT on expert demonstrations as a dynamically weighted auxiliary objective within the GRPO loss, enabling a smooth transition from off-policy imitation to on-policy exploration.

A second line of work addresses the instability that arises when optimizing against stale data. BAPO~\citep{xi2025bapo} observes that negative-advantage samples dominate the policy gradient when clipping bounds are fixed, blocking entropy-increasing updates; it dynamically adjusts clipping bounds to re-balance gradient contributions while preserving entropy. TOPR~\citep{roux2025tapered} takes a different approach, using tapered importance sampling to asymmetrically downweight unlikely negative trajectories while upweighting positive ones regardless of likelihood.

A third line of work strengthens the theoretical foundations of off-policy LLM training. \citet{mroueh2025revisiting} extend GRPO to an off-policy setting using samples from a lagged policy, proving that both the on-policy and off-policy variants provide a lower bound on policy improvement. AGRO~\citep{tang2025rl} introduces a generation consistency condition showing that the optimal policy satisfies relationships that hold across any sampling policy, yielding a unified on/off-policy algorithm with convergence guarantees.

Finally, ReMix~\citep{liang2025squeeze} presents a comprehensive framework that combines several of these ideas: a mixed-policy gradient with an increased update-to-data ratio, a KL-convex policy constraint that balances stability and flexibility, and policy reincarnation that transitions to on-policy training mid-way through optimization for stable convergence. A summary of motivations for the different off-policy approaches is provided in~\cref{tab:offpolicy}.

\begin{table}[!htb]
    \centering
    \begin{tabular}{l|l}
    \toprule
        \textbf{Method} & \textbf{Motivation}\\
        \midrule
        LUFFY~\citep{yan2025learning} & Leverage stronger models' reasoning trajectories \\
        CHORD~\citep{zhang2025policy} & Smoothly integrate SFT and RL training \\
        BAPO~\citep{xi2025bapo} & Stabilize off-policy optimization \\
        TOPR~\citep{roux2025tapered} & Enable stable learning from stale data \\
        Off-policy GRPO~\citep{mroueh2025revisiting} & Reduce rollout frequency and communication cost \\
        AGRO~\citep{tang2025rl} & Unify on-policy and off-policy learning \\
        ReMix~\citep{liang2025squeeze} & Reuse historical rollouts efficiently \\
        \bottomrule
    \end{tabular}
    \caption{Summary of off-policy RL methods for LLM training.}
    \label{tab:offpolicy}
\end{table}

\textbf{$\blacktriangle$ Experience Replay.}
Off-policy methods enable training on data generated by different policies, improving sample efficiency. A common technique is to maintain a replay buffer of past transitions and reuse them during training. When sampling from a replay buffer, two fundamental questions arise: (1)~what to replay, i.e, which trajectories should be selected; and (2)~when to replay, i.e, under what conditions replay should occur.

Experience replay has been extensively studied in classical RL. Hindsight experience replay~\citep{andrychowicz2017hindsight} addresses sparse rewards by relabeling failed trajectories with goals that were actually achieved, enabling learning from unsuccessful attempts. Prioritized experience replay~\citep{schaul2015prioritized,fujimoto2020equivalence} samples transitions proportionally to their temporal-difference error, focusing learning on experiences with higher learning potential. Other methods prioritize by state visitation frequency~\citep{sun2020attentive}, diversity~\citep{zhao2024efficient}, or retention strategies for continual learning~\citep{isele2018selective}.

Recent work has adapted these ideas to LLM training with domain-specific modifications. RRL~\citep{dou2025improving} uses a value model to identify promising intermediate states from early training and replays them dynamically as the policy's exploration capacity diminishes. BTP~\citep{chen2024enhancing} prioritizes samples based on P2Value (defined by output probability and test pass rate) to extract learning signal from partially correct programs in code generation. RLEP~\citep{zhang2025rlep} takes a simpler approach, mixing verified successful trajectories from earlier runs with fresh rollouts at every training update.

Other methods explore richer retrieval strategies. RePO~\citep{li2025repo} introduces multiple selection criteria: full-scope replay, recency-based selection of the most recent $K$ samples, reward-oriented sampling, and variance-driven selection to address vanishing gradients. EFRame~\citep{wang2025eframe} stores high-advantage samples from additional rollouts on difficult prompts and replays them periodically to maintain policy entropy. ARPO~\citep{lu2025arpo} targets sparse-reward settings in GUI agents, injecting a cached successful trajectory into GRPO training groups only when all sampled rollouts yield zero reward. A summary of these methods is provided in~\cref{tab:replay}.

\begin{table}[!htb]
    \centering
    \begin{tabular}{l|l|l}
    \toprule
        \textbf{Method} & \textbf{What to Replay} & \textbf{When to Replay}\\
        \midrule
        RRL~\citep{dou2025improving} & Value-identified promising states & After value model stabilizes \\
        BTP~\citep{chen2024enhancing} & High P2Value programs & Prioritized sampling during fine-tuning \\
        RLEP~\citep{zhang2025rlep} & Verified correct trajectories & Every update \\
        RePO~\citep{li2025repo} & Off-policy samples & After epoch $E_{\text{off}}$ \\
        EFRame~\citep{wang2025eframe} & High-advantage samples & Every $N$ steps \\
        ARPO~\citep{lu2025arpo} & Per-task successful trajectories & When group rewards are all zero \\
        \bottomrule
    \end{tabular}
    \caption{Summary of experience replay methods for LLM RL training.}
    \label{tab:replay}
\end{table}

\paragraph{Discussion.}
Current LLM RL is overwhelmingly on-policy: methods like GRPO~\citep{shao2024deepseekmath} and PPO~\citep{schulman2017proximal} sample fresh rollouts from the current policy, use them for a few gradient steps, and discard them. This is simple and avoids the distribution mismatch problems that make off-policy learning hard.

The downside is data efficiency: each trajectory is used for only a few gradient steps and then discarded. Unlike classical RL, where data collection requires costly environment interactions, LLM rollouts are simply forward passes through the model, making on-policy sampling relatively cheap. Still, each rollout consumes GPU compute, and the off-policy methods reviewed above aim to extract more learning signal per generated trajectory, whether by reusing demonstrations from stronger models~\citep{yan2025learning,zhang2025policy}, learning from stale rollouts more stably~\citep{xi2025bapo,roux2025tapered}, or replaying high-quality trajectories from earlier iterations~\citep{dou2025improving,li2025repo,lu2025arpo}. However, all of these methods still operate within the policy-gradient framework and rely on importance sampling to correct for the mismatch between old and current policies. Actor--critic methods with off-policy critics, which sidestep importance sampling entirely and are the standard off-policy approach in classical RL~\citep{haarnoja2018soft,fujimoto2018addressing}, remain unexplored for LLMs (see also~\cref{section:value_policy}).

A practical open question is how to decide which past trajectories are worth replaying. Current methods use task-specific heuristics~(\cref{tab:replay}); more general criteria based on estimated learning signal or policy divergence could improve robustness across tasks and training stages.

\begin{takeaway}
\textcolor{tabblue}{\textsf{Take-away}} \textbf{LLM RL is almost entirely on-policy}, which is simple but data-inefficient. Off-policy techniques can partially close this gap, though current approaches rely on importance sampling that often makes learning unstable or inefficient. The high-dimensional action space, sparse rewards, and long horizons in language model training make off-policy value estimation challenging and are likely explanations why current approaches primarily rely on on-policy methods.
\end{takeaway}

\subsection{Credit Assignment}
\label{section:credit_assignment}

Credit assignment is the problem of determining which actions are responsible for observed rewards~\citep{pignatelli2023survey}. A common inductive bias is temporal proximity: actions close in time to a reward are assumed to have caused it. This idea underlies temporal-difference (TD) learning, where value estimates are updated using the immediate reward plus a bootstrapped estimate of the next state's value. Q-learning and SARSA are canonical examples that perform such one-step backups. At the other end of the spectrum, Monte Carlo (MC) methods assign the full episodic return to every action in the trajectory. Early RL for LLMs used REINFORCE~\citep{williams1992simple}, which follows this approach, but the resulting gradient estimates suffer from high variance because every token in a response receives the same return signal regardless of its actual contribution.

More recent methods reduce this variance by using an advantage function, which measures how much better (or worse) an action is relative to the expected value under the current policy. PPO~\citep{schulman2017proximal} estimates token-level advantages via GAE~\citep{schulman2015high}, while GRPO~\citep{shao2024deepseekmath} estimates trajectory-level advantages by comparing rewards across multiple completions of the same prompt. Both yield lower-variance gradient estimates than raw returns, but differ in how they achieve this.

We organize credit assignment along two dimensions: (1)~backup depth, i.e, how far reward information is propagated through the sequence (one step in TD, the full episode in MC, or an interpolation such as TD($\lambda$)); and (2)~measure of action influence, i.e, what signal is used to determine how much credit each action receives and at what granularity (e.g., trajectory-level advantages in GRPO, token-level advantages in PPO, or refined variants targeting specific components of the policy gradient objective).

\paragraph{Backup depth.}
Most LLM training methods default to Monte Carlo updates: REINFORCE and GRPO compute advantages from complete trajectories and assign the same signal to every token. The main exception is PPO with GAE~\citep{schulman2015high}, which computes token-level advantages as an exponentially weighted sum of multi-step TD residuals, controlled by the discount factor $\gamma$ and the trace-decay parameter $\lambda$. GRPO-$\lambda$~\citep{parthasarathi2025grpo} brings a similar idea to the critic-free setting by reformulating eligibility traces using token-level log-probabilities, enabling finer-grained credit propagation within GRPO while preserving its lightweight design.

Recent work has sought to improve upon GRPO's uniform credit assignment by exploiting properties specific to the LLM setting. VinePPO~\citep{kazemnejad2024vineppo} exploits the ability to reset to any intermediate state by simply re-feeding the partial context to the model. This allows it to estimate the value of each state via Monte Carlo rollouts from that state, yielding unbiased, token-level advantages without a learned critic.

TEMPO~\citep{tran2025exploiting} exploits the tree structure that naturally emerges when sampling multiple responses per prompt. It constructs a prefix tree from grouped responses, computes nonparametric prefix values by aggregating descendant outcomes, and augments GRPO's group-relative signal with branch-gated TD corrections at divergence points, providing precise credit exactly where trajectories differ.

Backup depth also plays a role in the value-based and actor-critic methods discussed in~\cref{section:value_policy}. ShiQ~\citep{clavier2025shiq} uses $n$-step returns for Q-learning in LLMs, improving credit assignment through multi-step backups with provable convergence guarantees. TBRM~\citep{yuan2025trajectory} takes the opposite approach, optimizing a single TD error over the entire response rather than per-token errors. T-PPO~\citep{fan2025truncated} applies Extended GAE to estimate advantages from partial rollouts, enabling policy updates before full generation completes.

Overall, while backup depth is well studied in classical RL (spanning one-step TD, $n$-step returns, TD($\lambda$) and Monte Carlo), most LLM training methods default to Monte Carlo updates due to the sparse, outcome-based nature of typical rewards. The methods above can be viewed as attempts to recover finer-grained credit within this regime. Whether hybrid approaches that integrate bootstrapping more deeply into LLM training can outperform these Monte Carlo variants remains an open question.

\paragraph{Measure of action influence.}
Numerous recent algorithmic improvements focus on refining the GRPO~\citep{shao2024deepseekmath} objective. Using $x$ to denote the input prompt and $y_i = (y_{i,1}, \ldots, y_{i,|y_i|})$ to denote the $i$-th sampled response (corresponding to the state-action trajectory in~\cref{eq:policy_gradient}), GRPO optimizes the following:
\begin{equation}
\mathcal{J}_{\text{GRPO}}(\theta) = \mathbb{E}_{x \sim \mathcal{D}, \{y_i\}_{i=1}^G \sim \pi_{\theta_{\text{old}}}(\cdot|x)} 
\left[ \frac{1}{G} \sum_{i=1}^G \frac{1}{|y_i|} \sum_{t=1}^{|y_i|} \min \left( w_{i,t}(\theta) \hat{A}_{i,t}, \, \text{clip}\left( w_{i,t}(\theta), 1 - \epsilon, 1 + \epsilon \right) \hat{A}_{i,t} \right) \right],
\end{equation}

where $G$ is the number of generated responses to each query $x$ (i.e, the group size), and the importance ratio $w_{i,t}(\theta)$ and advantage $\hat{A}_{i,t}$ of token $y_{i,t}$ are:

\[
\textcolor{tabred}{w_{i,t}(\theta) = \frac{\pi_\theta(y_{i,t} | x, y_{i,<t})}{\pi_{\theta_{\text{old}}}(y_{i,t} | x, y_{i,<t})}}, 
\qquad 
\textcolor{tabblue}{\hat{A}_{i,t} = \hat{A}_i = \frac{r(x, y_i) - \text{mean}\left(\{r(x, y_j)\}_{j=1}^G\right)}{\text{std}\left(\{r(x, y_j)\}_{j=1}^G\right)}},
\]

respectively, where all the tokens in $y_i$ share the same advantage $\hat{A}_i$. The clip function is defined as:

\[
\textcolor{taborange}{\text{clip}(w_{i,t}(\theta),\, 1-\epsilon,\, 1+\epsilon) = \max\left(1-\epsilon,\, \min\left(w_{i,t}(\theta),\, 1+\epsilon\right)\right)}
\]

Recent improvements primarily target three components of this objective: the precision of the \textcolor{tabblue}{advantage estimation}, the stability of the \textcolor{tabred}{importance sampling} mechanism, and problems introduced by the \textcolor{taborange}{clipping} function. An overview of these methods is provided in~\cref{tab:op_checked}.

\textbf{Advantage estimation.}
A central limitation of GRPO is that all tokens in a response share the same advantage, making the method unable to distinguish which tokens actually contributed to the outcome. Several methods address this at different granularities. At the token level, \textcolor{tabblue}{VinePPO}~\citep{kazemnejad2024vineppo} generates auxiliary Monte Carlo rollouts from intermediate states to obtain per-token value estimates, while \textcolor{tabblue}{TPPO}~\citep{fan2025truncated} introduces Extended Generalized Advantage Estimation (EGAE) to enable optimization over partially generated responses. \textcolor{tabblue}{GFPO}~\citep{shrivastava2025sample} takes a different approach, sampling larger groups and filtering for token efficiency before computing advantages. Even with improved estimates, reward signals for the same token can conflict across completions within a group; \textcolor{tabblue}{GTPO}~\citep{simoni2025gtpo} addresses this by masking conflicting token gradients to prevent optimization collapse. Also operating at the token level, \textcolor{tabblue}{GMPO}~\citep{zhao2025geometric} stabilizes training by applying a geometric mean over token-level losses instead of the arithmetic mean, dampening the influence of outlier signals.

Beyond token-level refinements, several methods rethink the group structure itself. Group-based methods estimate advantages from independent response samples, which redundantly recompute shared prefixes and limit exploration. \textcolor{tabblue}{TreePO}~\citep{li2025treepo} mitigates this via tree-structured search sampling with dynamic branching and early pruning, improving exploration and training stability. \textcolor{tabblue}{SPO}~\citep{xu2025single} eliminates the group structure entirely, replacing it with a persistent Bayesian value tracker as a per-prompt baseline and global advantage normalization, avoiding the zero-gradient problem that arises when all responses in a group receive the same reward. For multi-turn agent settings, where credit must be assigned across interaction steps rather than just tokens, \textcolor{tabblue}{GiGPO}~\citep{feng2025group} introduces a two-level grouping structure: episode-level advantages capture overall trajectory quality, while step-level advantages are computed by grouping actions taken from the same environment state across trajectories, enabling fine-grained credit without a learned critic.

Beyond the objective itself, several methods revisit the normalization strategy underlying GRPO. The original formulation~\citep{shao2024deepseekmath} uses group-relative normalization, but \textcolor{tabblue}{Dr.\ GRPO}~\citep{liu2025understanding} argues that this introduces length and difficulty biases, proposing normalization by a global constant instead. \textcolor{tabblue}{BNPO}~\citep{xiao2025bnpo} introduces further flexibility by replacing static normalization with an adaptive beta distribution, while for settings with multiple reward signals, \textcolor{tabblue}{GDPO}~\citep{liu2026gdpo} decouples normalization across each reward independently, ensuring stable updates across heterogeneous metrics.

\textbf{Importance sampling.}
The importance sampling (IS) term corrects for the distribution mismatch between the policy used for data collection and the policy being updated. \textcolor{tabred}{GSPO}~\citep{zheng2025group} identifies a mismatch between sequence-level rewards and token-level ratios, and proposes a length-normalized sequence-level IS approach. To maintain training signals during late-stage convergence, \textcolor{tabred}{AAPO}~\citep{xiong2025aapo} introduces momentum into the IS term. In contrast, \textcolor{tabred}{GPG}~\citep{chu2025gpg} takes a minimalist path by removing IS and normalization entirely, directly optimizing the policy gradient objective via cross-entropy. \textcolor{tabred}{GEPO}~\citep{zhang2025group} tackles the challenges of decentralized training latency by computing group-level expectations for IS weights, while in multi-turn scenarios, \textcolor{tabred}{ATPO}~\citep{zong20262} further refines this with tree-structured, turn-level importance sampling for hierarchical tasks. Rather than correcting off-policy staleness through IS, \textcolor{tabred}{OPO}~\citep{hao2025policy} sidesteps the problem entirely by employing on-policy single-update regimes.

\textbf{Clipping.}
The hard clipping mechanism in the GRPO objective creates an all-or-nothing gate: gradients are either fully preserved or completely zeroed out. \textcolor{taborange}{SAPO}~\citep{gao2025soft} softens this boundary with a sigmoid-shaped clipping function that allows gradients on negative tokens to decay more smoothly. \textcolor{taborange}{GPPO}~\citep{su2025klear} goes further by incorporating gradient signals from samples beyond the clip boundary while constraining these out-of-bound signals within a defined range. \textcolor{taborange}{DPPO}~\citep{qi2026rethinking} replaces clipping altogether with divergence-based masking, constraining updates to remain within the trust region. \textcolor{taborange}{DAPO}~\citep{yu2025dapo} takes yet another angle, observing that the small upper bound of the clip range limits exploratory behavior, and proposes raising it to encourage greater exploration.

\begin{table}[!hbt]
    \centering
    \footnotesize
    \begin{tabular}{lp{6.8cm}p{6.8cm}}
        \toprule
        \textbf{Method} & \textbf{Problem} & \textbf{Solution} \\
        \midrule
        \multicolumn{3}{l}{\scriptsize\textcolor{tabblue}{$\blacksquare$}~Advantage Estimation \quad \textcolor{tabred}{$\blacksquare$}~Importance Sampling \quad \textcolor{taborange}{$\blacksquare$}~Clipping} \\
        \midrule
        \multicolumn{3}{c}{\textbf{2024}} \\
        \midrule
        \textcolor{tabblue}{VinePPO} & All tokens share the same trajectory-level advantage; no per-token credit. & Reset to intermediate states via context re-feeding; estimate per-token $V(s_t)$ via MC rollouts from each state.\\
        \midrule
        \multicolumn{3}{c}{\textbf{2025}} \\
        \midrule
        \textcolor{tabblue}{BNPO} & Group-relative normalization uses fixed mean/std, which is brittle when reward distributions are skewed. & Fit a Beta distribution to group rewards and normalize advantages adaptively.\\
        \textcolor{tabblue}{Dr.\ GRPO} & Dividing by per-group std biases toward short or easy prompts with low reward variance. & Remove per-group std normalization; divide by a global constant proportional to response length.\\
        \textcolor{tabblue}{GFPO} & Verbose filler tokens inflate per-token reward, rewarding length over quality. & Sample $G' > G$ responses per prompt; retain only the top-$k$ by token efficiency for advantage computation. \\
        \textcolor{tabblue}{GiGPO} & Trajectory-level advantage cannot distinguish which interaction steps caused success or failure in multi-turn tasks. & Two-level grouping: episode-level advantages from full trajectories, step-level advantages from actions taken at shared environment states across trajectories.\\
        \textcolor{tabblue}{GMPO} & Arithmetic mean over token losses lets a few outlier tokens dominate the gradient. & Replace arithmetic mean with geometric mean over token-level losses, dampening outlier influence.\\
        \textcolor{tabblue}{GTPO} & The same token can appear in both positive- and negative-advantage responses, producing conflicting gradients. & Mask tokens whose gradient signs conflict across responses within the group before updating. \\
        \textcolor{tabblue}{TPPO} & Policy must wait for full sequence generation before computing advantages; wastes GPU time on long responses. & Extended GAE (EGAE) estimates advantages from partial rollouts, enabling updates before generation completes.\\
        \textcolor{tabblue}{TreePO} & Independent response sampling redundantly generates shared prefixes and limits exploration diversity. & Tree-structured sampling with dynamic branching at divergence points and early pruning of low-reward branches. \\
        \textcolor{tabblue}{SPO} & All-correct or all-incorrect groups produce zero advantage (degenerate groups); group synchronization bottlenecks in distributed training. & Replace group baseline with a persistent Bayesian value tracker per prompt; apply global advantage normalization across prompts. \\
        \textcolor{tabred}{AAPO} & As training converges, within-group reward variance vanishes, producing near-zero advantages and stalled learning. & Add momentum to the advantage by incorporating the gap between current policy and reference policy log-probabilities.\\
        \textcolor{tabred}{GEPO} & Asynchronous distributed training causes stale $\pi_{\theta_{\text{old}}}$, leading to exploding IS ratios. & Smooth IS weights by computing group-level expectations over the ratio distribution.\\
        \textcolor{tabred}{GPG} & Surrogate loss, IS correction, and KL penalty add complexity with marginal benefit over direct optimization. & Remove IS ratios and KL penalty; optimize the policy gradient objective directly via weighted cross-entropy.\\
        \textcolor{tabred}{GSPO} & Token-level IS ratios mismatch with sequence-level rewards; product of many per-token ratios has high variance. & Compute a single sequence-level IS ratio, normalized by response length, with sequence-level clipping.\\
        \textcolor{tabred}{OPO} & Multiple gradient steps per batch cause the policy to drift from $\pi_{\theta_{\text{old}}}$, making IS corrections inaccurate. & Perform a single gradient step per batch (on-policy); use a length-weighted baseline instead of IS correction.\\
        \textcolor{taborange}{DAPO} & Symmetric clip range $[1{-}\epsilon, 1{+}\epsilon]$ with small $\epsilon$ suppresses exploration; entropy collapses. & Raise the upper clip bound (Clip-Higher) to allow entropy-increasing updates; add dynamic sampling and overlong reward shaping.\\
        \textcolor{taborange}{GPPO} & Hard clipping zeroes out gradients for high-entropy reflective tokens that exceed the clip boundary. & Allow gradient signals from beyond the clip boundary, constrained within a secondary range.\\
        \textcolor{taborange}{SAPO} & Hard clipping creates a discontinuous gradient landscape, destabilizing training. & Replace hard clip with a sigmoid-shaped soft clipping function for smooth gradient decay. \\
        \midrule
        \multicolumn{3}{c}{\textbf{2026}} \\
        \midrule
        \textcolor{tabblue}{GDPO} & With multiple reward signals, group normalization mixes heterogeneous reward scales, making groups indistinguishable. & Decouple normalization: compute group-relative advantages independently for each reward signal.\\
        \textcolor{tabred}{ATPO} & Sequence-level IS is too coarse for multi-turn tasks where individual turns have different staleness. & Compute IS ratios and advantages at the turn level using a tree-structured decomposition of the trajectory.\\
        \textcolor{taborange}{DPPO} & Ratio clipping is a loose proxy for policy divergence; fails for long-tailed token distributions. & Replace clipping with a direct divergence constraint that masks tokens exceeding a per-token KL threshold.\\
        \bottomrule
    \end{tabular}
    \caption{Summary of GRPO variants, identified problems, and their solutions.}
    \label{tab:op_checked}
\end{table}

\paragraph{Discussion.}
Credit assignment in LLM RL currently relies on two simplifying defaults: Monte Carlo backup (propagating a single outcome reward uniformly to all tokens) and trajectory-level advantage estimation (assigning the same advantage to every token via GRPO-style group normalization). These defaults are simple and work well for short-answer tasks with verifiable rewards~\citep{guo2025deepseek,shao2024deepseekmath}, but they become limiting as responses grow longer and rewards become sparser.

On backup depth, most methods remain purely Monte Carlo because sparse outcome rewards provide no per-step signal to bootstrap from. The few exceptions, such as GRPO-$\lambda$~\citep{parthasarathi2025grpo} (eligibility traces), TEMPO~\citep{tran2025exploiting} (branch-gated TD corrections), and VinePPO~\citep{kazemnejad2024vineppo} (MC rollouts from intermediate states), show consistent improvements by introducing finer temporal structure. Whether process reward models or learned critics can supply per-step signals that enable deeper bootstrapping at scale remains an open question.

On the measure of action influence, the GRPO variants in~\cref{tab:op_checked} refine three interacting components of the policy gradient objective: advantage estimation, importance sampling, and clipping. These components are coupled in practice, yet most methods modify one component while holding the others at GRPO's defaults, and systematic studies of how these three components interact remain scarce.

\begin{takeaway}
\textcolor{tabblue}{\textsf{Take-away}} \textbf{Credit assignment for LLMs may benefit from looking beyond incremental GRPO refinements}: most current work targets individual components (advantage estimation, IS, or clipping) in isolation, but these components interact, and the underlying MC backup with trajectory-level credit remains unchanged. More fundamental advances, such as integrating bootstrapping into LLM training, developing scalable per-step reward signals, or rethinking the policy gradient objective itself, may yield larger gains than continued refinement within the GRPO framework.
\end{takeaway}

\section{Other RL Frameworks in LLM Training}
\label{section:other_rl}

The RL methods discussed so far have largely focused on single-agent, online settings. However, the broader RL landscape encompasses a rich variety of frameworks that address different structural assumptions and learning paradigms, many of which have found productive applications in LLM training. In this section, we briefly overview four such frameworks: meta-RL, multi-agent RL (MARL), offline RL, and hierarchical RL (HRL). Our coverage is necessarily selective, highlighting representative works to illustrate how each framework has been applied to LLM training rather than providing exhaustive reviews.

\paragraph{Meta-RL}
Meta-RL~\citep{beck2023survey} refers to a framework in which an agent learns to rapidly adapt to new tasks by leveraging experience across a distribution of prior tasks, rather than training a separate policy for each one.
Recent work has drawn connections between meta-RL and LLM reasoning. \citet{qu2025optimizing} reframe long chain-of-thought reasoning as a meta-RL problem, where each reasoning block constitutes an episode in an unknown MDP. They minimize cumulative regret by augmenting the outcome reward with a progress bonus that measures the information gain of each reasoning step toward eventual success.
In interactive agentic settings, LAMER~\citep{jiang2025meta} trains LLM agents with cross-episode RL, where the agent learns to explore strategically in early episodes and exploit gathered information in later ones, mirroring the explore-then-exploit structure central to meta-RL. ORBIT~\citep{lin2026scaling} extends this cross-episode idea by training across a diverse distribution of tasks, aiming for generalization to entirely unseen environments at test time.

\paragraph{MARL}
In the MARL setting~\citep{huh2023multi}, multiple agents must learn to coordinate their behaviors, through communication~\citep{zhu2024survey}, cooperation, or competition, to achieve collective goals. MARL algorithms such as multi-agent PPO~\citep{yu2022surprising} have been applied to the training of debating LLM agents~\citep{park2025maporl}. Meanwhile, several works extend the GRPO objective to multi-agent settings. \citet{liu2025llm} formalize LLM collaboration as a Dec-POMDP~\citep{bernstein2002complexity} and train LLMs for collaborative writing, while AT-GRPO~\citep{zhao2025stronger} introduces granular credit assignment across agents and turns within a shared task. Beyond policy optimization, MPDF~\citep{yang2025learning} trains a decentralized meta-policy over deliberative actions (e.g., persist, refine, or concede), enabling agents to adaptively coordinate based on internal confidence and group consensus. Finally, MARFT~\citep{liao2025marft} provides a theoretical framework bridging classical MARL theory and LLM-based multi-agent systems, addressing challenges unique to the LLM setting such as asynchronous execution and heterogeneous agent roles.

\paragraph{Offline RL}
Offline RL~\citep{levine2020offline} learns policies entirely from pre-collected datasets, avoiding the computational expense of generating new samples during training. In the LLM context,s this typically means learning from previously collected datasets of human feedback or demonstrations.
The most prominent example is DPO~\citep{rafailov2023direct}. Despite using a supervised classification loss, DPO is mathematically equivalent to solving the standard KL-regularized reward maximization objective under a closed-form optimal policy. While DPO operates on pairwise preference data, such annotations can be expensive to collect at scale. DRO~\citep{richemond2024offline} relaxes this requirement by learning from single-trajectory datasets consisting of a prompt, a response, and scalar feedback (e.g., a thumbs-up or thumbs-down), which are more naturally abundant in real-world applications. A separate limitation of DPO is that it treats all tokens uniformly and operates at the response level, making it poorly suited for credit assignment in multi-step reasoning tasks with sparse rewards. OREO~\citep{wang2024offline} addresses this by jointly learning a policy and an explicit value function via the soft Bellman equation, enabling token-level credit assignment without requiring pairwise data. We refer the reader to~\citet{xiao2024comprehensive} for a comprehensive survey of DPO and its many variants.

Beyond developing new offline RL algorithms, several studies have investigated when and how well offline methods perform relative to their online counterparts.
\citet{tian2025exploring} systematically evaluate offline methods, specifically DPO and its length-desensitized variant LD-DPO~\citep{liu2024length}, on reasoning benchmarks where online RL methods have traditionally dominated, and find that these simpler offline approaches can still yield meaningful improvements.
In contrast, \citet{lanchantin2025bridging} compare offline DPO against online GRPO across a spectrum of training regimes (offline, semi-online, and fully online) on both verifiable and non-verifiable tasks, and find that online and semi-online methods consistently and substantially outperform purely offline training.
Rather than training the LLM policy itself, PNLC~\citep{hong2025planning} takes a different approach: it uses offline RL with an IQL-style~\citep{kostrikov2021offline} objective to train a lightweight, goal-conditioned value function, which is then used at inference time to evaluate and refine the LLM's reasoning in multi-turn interactive tasks such as negotiation and tool use.

\paragraph{HRL}
HRL~\citep{pateria2021hierarchical} decomposes complex tasks into high-level goals and low-level actions, where higher levels handle subgoal selection and long-horizon planning while lower levels execute primitive actions. Recent work reveals that such hierarchical structure can emerge naturally in RL-trained LLMs. \citet{wang2025emergent} show that RL training induces a two-phase dynamic, first consolidating low-level procedural skills, then shifting to high-level strategic planning, and propose HICRA, which concentrates optimization on planning tokens rather than applying it uniformly as in standard GRPO. Complementarily, \citet{tan2025bottom} analyze hierarchy within the model architecture itself, finding that early layers tend to explore while top layers refine, and propose BuPO, which first optimizes an intermediate-layer policy before fine-tuning the full model.
Other works exploit hierarchical structure through reward design and inference-time decomposition. L2T~\citep{wang2025learning} segments reasoning chains into episodes with information-theoretic dense rewards, providing fine-grained learning signals. Cog-Rethinker~\citep{sun2025cog} introduces hierarchical metacognitive rollouts that progressively refine failed solutions at increasing levels of abstraction. ArCHer~\citep{zhou2024archer} combines utterance-level value functions with token-level policy optimization, enabling hierarchical credit assignment in multi-turn dialogue tasks.

\medskip
\noindent The frameworks surveyed above represent only a slice of the growing intersection between RL and LLM training. Each sub-area is developing rapidly, and we believe dedicated surveys for meta-RL, MARL, offline RL, and HRL in the LLM context will be valuable future contributions.

\section{Conclusion}
This survey has examined RL techniques for LLMs from a first-principles perspective, organized around the design decisions involved in building an RL algorithm from the ground up. We began with MDP creation, examining how reward functions, state spaces, action spaces, termination conditions, and discount factors are defined for LLMs. We then reviewed exploration mechanisms, including temperature sampling, entropy regularization, intrinsic motivation, tree search, and curriculum learning. Finally, we categorized learning methods along four classical RL dimensions: model-free versus model-based, value-based versus policy-based versus actor-critic, on-policy versus off-policy, and credit assignment.

This modular, RL-centric organization reveals a non-uniform distribution of research effort. Some categories, such as critic-free policy gradients and Monte Carlo credit assignment, are densely populated, while others, such as value-based methods, off-policy actor-critic training, and bootstrapping-based credit assignment, remain largely unexplored despite well-established counterparts in classical RL. By making these gaps explicit, our taxonomy complements existing LLM-centric surveys, which naturally emphasize areas where research is already concentrated.

We hope this perspective serves as a useful reference for researchers in both RL and LLMs. The structural gaps identified here do not necessarily point to easy wins, as there are often good reasons why certain classical RL techniques have not yet been adopted in LLM training. Nevertheless, we believe they represent directions worth investigating, and that a shared conceptual framework between the two communities is a productive starting point for that work.

\bibliographystyle{ACM-Reference-Format}
\bibliography{sample-base}

\end{document}